\documentclass[pdflatex,sn-mathphys-num]{sn-jnl}% Math and Physical Sciences Numbered Reference Style
\usepackage{graphicx}%
\usepackage{multirow}%
\usepackage{amsmath,amssymb,amsfonts}%
\usepackage{amsthm}%
\usepackage{mathrsfs}%
\usepackage[title]{appendix}%
\usepackage{xcolor}%
\usepackage{textcomp}%
\usepackage{manyfoot}%
\usepackage{booktabs}%
\usepackage{algorithm}%
\usepackage{algorithmicx}%
\usepackage{algpseudocode}%
\usepackage{listings}%
\theoremstyle{thmstyleone}%
\newtheorem{theorem}{Theorem}%  meant for continuous numbers
\newtheorem{proposition}[theorem]{Proposition}% 
\newtheorem{corollary}[theorem]{Corollary}
\theoremstyle{thmstyletwo}%

\theoremstyle{thmstylethree}%
\newtheorem{lemma}[theorem]{Lemma} 
\definecolor{reviewblue}{RGB}{0, 70, 140}
\definecolor{reviewred}{RGB}{180, 30, 40}

\begin{document}

\title[Robust and Fast MMD Variance Estimation]{Finite-Sample Unbiased Variance of MMD under Unbalanced Sampling: Exact Estimation and Quasi-Linear Computation}


\author[1,2]{\fnm{Shijie} \sur{Zhong}}\email{zhongsj@mail.nwpu.edu.cn}

\author[1,2]{\fnm{Yikun} \sur{Yang}}\email{yikunyang@mail.nwpu.edu.cn}

\author[3]{\fnm{Da} \sur{Gong}}\email{641636405@qq.com}

\author*[1,2,4]{\fnm{Jiangfeng} \sur{Fu}}\email{fjf@nwpu.edu.cn}

\affil*[1]{\orgdiv{School of Power and Energy}, \orgname{Northwestern Polytechnical University}, \orgaddress{\street{1 Dongxiang Road}, \city{Xi'an}, \postcode{710129}, \state{Shanxi}, \country{China}}}

\affil[2]{\orgdiv{Advanced Power Research Institute}, \orgname{Northwestern Polytechnical University}, \orgaddress{\street{619 Jicui Road}, \city{Chengdu}, \postcode{610299}, \state{Sichuan}, \country{China}}}

\affil[3]{\orgname{Taihang Laboratory}, \orgaddress{\city{Chengdu}, \postcode{610200}, \state{Sichuan}, \country{China}}}

\affil[4]{\orgdiv{Science and Technology on Altitude Simulation Laboratory}, \orgname{AECC Sichuan GasTurbine Establishment}, \orgaddress{\city{Mianyang}, \postcode{621703}, \state{Sichuan}, \country{China}}}

\abstract{
	Accurately and efficiently estimating the variance of the Maximum Mean Discrepancy (MMD) remains challenging, particularly for unbalanced sample sizes. In this paper, we derive a finite-sample unbiased estimator of the MMD variance. To overcome the traditional $\mathcal{O}(N^2)$ computational bottleneck, we develop a recursive prefix-suffix accumulation scheme for the Laplace kernel, reducing the computational complexity to $\mathcal{O}(N \log N)$ while requiring $\mathcal{O}(N)$ memory. Experimental results verify the theoretical exactness and numerical stability of the proposed estimator and demonstrate its scalability on large datasets. Furthermore, the method proves effective for monitoring distributional convergence during the training of Time-series Generative Adversarial Networks (TimeGAN).
}



\maketitle

\section{Introduction}
\label{sec1}
The fundamental problem of nonparametric two-sample testing is to detect differences between two probability distributions given access only to samples drawn from them. Maximum Mean Discrepancy (MMD) is a kernel-based nonparametric statistic proposed for this task \cite{borgwardt2006integrating, gretton2012kernel}. When equipped with a characteristic kernel, MMD defines a metric on the space of probability distributions, and thus equals zero if and only if the two distributions are identical \cite{muandet2017kernel, sriperumbudur2010hilbert}. It transforms the distance measure between any two distributions into a linear operation in the Hilbert space defined by a kernel function. In particular, MMD has been successfully applied to tasks such as domain adaptation \cite{pan2010domain, muandet2013one, li2018domain} and generative modeling \cite{gretton2012optimal, liu2020learning,deka2023mmd}, and shows promising potential in Bayesian model calibration \cite{fukumizu2011kernel, wei2024probabilistic, wei2025bayesian}.

The reliability of MMD depends crucially on accurately characterizing its variance. When the two samples are balanced, the MMD is classically framed as a standard U-statistic, allowing its variance properties to be readily established \cite{sutherland2019unbiased}. However, practical applications frequently involve unbalanced samples. To accommodate this asymmetry, Wei et al. \cite{wei2025maximum} constructed an expanded function to cast the MMD as a generalized U-statistic \cite{kim2022minimax, schrab2023mmd}, deriving an exact variance expression formulated through kernel mean embedding and Hilbert--Schmidt operators. Motivated by the balanced-sample analysis in \cite{sutherland2019unbiased}, an exact and unbiased variance estimator for the generalized MMD in \cite{wei2025maximum} can likewise be constructed.

Nevertheless, the exact variance estimator remains computationally expensive, as its quadratic complexity imposes substantial burdens on both memory usage and runtime in large-scale applications. Existing acceleration strategies exhibit clear trade-offs: global approximation methods, such as subsampling or random feature projections, scale well to high-dimensional data but introduce inherent approximation errors \cite{zaremba2013b, zhao2015fastmmd}. Analytical acceleration offers an exact alternative by restructuring independent summations into ordered linear recursions \cite{adelsonvelskii1962algorithm, johnson1978selecting}, a strategy previously applied in classical statistics for measures like Kendall's tau and distance covariance to reduce complexity from $\mathcal{O}(N^2)$ to $\mathcal{O}(N \log N)$ \cite{knight1966computer, huo2016fast}. In the MMD context, Bodenham and Kawahara \cite{bodenham2023eummd} leveraged the semigroup property of the Laplace kernel to achieve a quasi-linear-time algorithm, referred to as efficient computation of univariate MMD (euMMD), in univariate settings.

However, their framework was specifically designed for evaluating the mean discrepancy and does not extend to unbiased variance estimation, which requires richer structural information, including localized interaction summaries. To overcome these limitations, we propose a generalized recursive accumulation framework based on the euMMD framework.

The remainder of this paper is organized as follows. Section~\ref{sec2} reviews the background on MMD and existing euMMD techniques. Section~\ref{sec3} presents a finite-sample unbiased estimator for the MMD variance and a generalized quasi-linear $\mathcal{O}(N \log N)$ accelerated algorithm. Section~\ref{sec4} provides numerical experiments, including theoretical correctness and computational efficiency of the proposed method. In particular, we compare with FastMMD, which employs a Random Fourier Features (RFF) approximation (detailed in Appendix~C), and evaluate its performance in the training of Time-series Generative Adversarial Networks (TimeGAN).

\section{Background}
\label{sec2}

\subsection{Maximum Mean Discrepancy}

Assume $X, X' \sim \mathbb{P}$ and $Y, Y' \sim \mathbb{Q}$ are independent copies. The maximum mean discrepancy is defined as
\begin{equation}
	\mathrm{MMD}^2(\mathbb{P}, \mathbb{Q}) = \mathbb{E}_{X, X'} [k(X, X')] + \mathbb{E}_{Y, Y'} [k(Y, Y')] - 2\mathbb{E}_{X, Y} [k(X, Y)],
	\label{MMD}
\end{equation}
where $\mathbb{E}_{X, X'}$ denotes the expectation with respect to the independent random variables $X, X' \sim \mathbb{P}$ (with $\mathbb{E}_{Y, Y'}$ and $\mathbb{E}_{X, Y}$ defined analogously), and $k(\cdot, \cdot)$ is a symmetric positive definite kernel function.

The effectiveness of MMD relies on characteristic kernels, under which the kernel mean embedding uniquely represents probability distributions in the associated reproducing kernel Hilbert space (RKHS) \cite{fukumizu2008characteristic}. In this case, MMD defines a proper metric on distributions, ensuring that any difference between them is detectable (i.e., $\mathrm{MMD}(\mathbb{P}, \mathbb{Q}) = 0$ if and only if $\mathbb{P} = \mathbb{Q}$) \cite{fukumizu2004dimensionality,fukumizu2007kernel}.

In practice, translation-invariant kernels are particularly appealing due to their simple analytic form and favorable computational properties, as characterized by Bochner's theorem. Two commonly used examples are the Gaussian kernel and the Laplace kernel:
\begin{align}
	k_\mathrm{G}(\mathbf{x}, \mathbf{x}'; \sigma) &= \exp \left( - \frac{\|\mathbf{x} - \mathbf{x}'\|^2}{2\sigma^2} \right), \label{eq:gaussian} \\
	k_\mathrm{L}(\mathbf{x}, \mathbf{x}'; \sigma) &= \exp \left( - \frac{\|\mathbf{x} - \mathbf{x}'\|}{\sigma} \right), \label{eq:laplace}
\end{align}
where the kernel parameter $\sigma$ controls the scale of the feature space. A common choice is the median heuristic \cite{gretton2005kernel}, where $\sigma$ is set to the median of all pairwise sample distances.

Given a chosen kernel and i.i.d. finite samples $\mathbf{X} = \{\mathbf{x}_i\}_{i=1}^n$ and $\mathbf{Y} = \{\mathbf{y}_j\}_{j=1}^m$ drawn from $\mathbb{P}$ and $\mathbb{Q}$, respectively, an unbiased estimator of the MMD can be computed as
\begin{equation}
	\widehat{\mathrm{MMD}}^2(\mathbf{X}, \mathbf{Y})
	= \frac{1}{n(n-1)} \sum_{i \neq i'} k(\mathbf{x}_i, \mathbf{x}_{i'})
	+ \frac{1}{m(m-1)} \sum_{j \neq j'} k(\mathbf{y}_j, \mathbf{y}_{j'}) 
	- \frac{2}{nm} \sum_{i,j} k(\mathbf{x}_i, \mathbf{y}_j).
	\label{uMMD}
\end{equation}

The estimator can be written in matrix form as
\begin{equation}
	\widehat{\mathrm{MMD}}^2(\mathbf{X},\mathbf{Y}) = \frac{1}{(n)_2} \mathbf{1}_n^\top \tilde{\mathbf{K}}_{\mathbf{XX}} \mathbf{1}_n + \frac{1}{(m)_2} \mathbf{1}_m^\top \tilde{\mathbf{K}}_{\mathbf{YY}} \mathbf{1}_m - \frac{2}{nm} \mathbf{1}_n^\top \mathbf{K}_{\mathbf{XY}} \mathbf{1}_m.
	\label{matrixMMD}
\end{equation}
where $\mathbf{K}_{\mathbf{XX}} \in \mathbb{R}^{n\times n}$ and $\mathbf{K}_{\mathbf{YY}} \in \mathbb{R}^{m\times m}$ denote the within-sample Gram matrices, and $\mathbf{K}_{\mathbf{XY}} \in \mathbb{R}^{n\times m}$ denotes the cross-sample kernel matrix with entries $(\mathbf{K}_{\mathbf{XY}})_{ij}=k(x_i,y_j)$. Setting the diagonal entries of $\mathbf{K}_{\mathbf{XX}}$ and $\mathbf{K}_{\mathbf{YY}}$ to zero yields $\tilde{\mathbf{K}}_{\mathbf{XX}}$ and $\tilde{\mathbf{K}}_{\mathbf{YY}}$. Furthermore, $\mathbf{1}_n$ and $\mathbf{1}_m$ denote vectors of ones with dimensions $n$ and $m$, and $(n)_k = {n!}\big/{(n-k)!}$ denotes the falling factorial.

\subsection{Variance of the unbiased MMD estimator}
\label{sec:variance_background}

The variance of the MMD estimator plays an important role in its inferential behavior. We introduce kernel mean embeddings and Hilbert-Schmidt operators. For a distribution $\mathbb{P}$, the mean embedding is defined as
\begin{equation}
	\mu_{\mathbb{P}} = \mathbb{E}_{X \sim \mathbb{P}}[k(X, \cdot)] \in \mathcal{H},
\end{equation}
and the centered covariance operator is defined as
\begin{equation}
	C_{\mathbb{P}}
	=
	\mathbb{E}_{X \sim \mathbb{P}}
	\big[
	(k(X,\cdot) - \mu_{\mathbb{P}}) \otimes (k(X,\cdot) - \mu_{\mathbb{P}})
	\big],
	\label{CP}
\end{equation}
where $\otimes$ denotes the tensor product.

For balanced samples ($n=m$), the MMD is reduces to a standard second-order U-statistic with a known closed-form variance (Eq.~4 in \cite{wei2025maximum}):
\begin{equation}
	\mathrm{Var}(\widehat{\mathrm{MMD}}^2) =\frac{4}{n}\langle \Delta\mu, \left(C_{\mathbb{P}} + C_{\mathbb{Q}}\right) \Delta\mu  \rangle_{\mathcal{H}} + \frac{2}{n(n-1)}\|C_{\mathbb{P}}+C_{\mathbb{Q}}\|_{\mathrm{HS}}^2,
\end{equation}
where $\Delta\mu = \mu_{\mathbb{P}} - \mu_{\mathbb{Q}}$.

However, practical scenarios frequently involve unbalanced samples ($n \neq m$), precluding independent pairing. In this regime, MMD is formally cast as a generalized two-sample U-statistic, utilizing an expanded four-argument symmetric kernel:
\begin{equation}
	h(\mathbf{x}_i, \mathbf{x}_{i'}; \mathbf{y}_j, \mathbf{y}_{j'})
	=
	k(\mathbf{x}_i, \mathbf{x}_{i'}) + k(\mathbf{y}_j, \mathbf{y}_{j'})
	- \frac{1}{2}
	\sum_{a \in \{i,i'\}} \sum_{b \in \{j,j'\}}
	k(\mathbf{x}_a, \mathbf{y}_b).
\end{equation}

According to the theory of generalized U-statistics, the exact variance can be expressed as a combinatorial sum of conditional variance components\cite{sen1974weak}. A direct finite-sample expansion was recently proposed by Wei et al. \cite{wei2025maximum}, but appears to contain an inconsistency in the cross-covariance projection terms, resulting in a more complicated formulation. We revisit the combinatorial derivation in Appendix~\ref{App1} and obtain the corresponding variance formula:
\begin{align}
	V:=\mathrm{Var}(\widehat{\mathrm{MMD}}^2) &=\frac{4}{n}\langle \Delta\mu, C_{\mathbb{P}} \Delta\mu \rangle_{\mathcal{H}} + \frac{4}{m}\langle \Delta\mu, C_{\mathbb{Q}} \Delta\mu \rangle_{\mathcal{H}}\nonumber \\
	& +\frac{2}{n(n-1)}\|C_{\mathbb{P}}\|_{\mathrm{HS}}^2 + \frac{2}{m(m-1)}\|C_{\mathbb{Q}}\|_{\mathrm{HS}}^2 + \frac{4}{nm}\langle C_{\mathbb{P}}, C_{\mathbb{Q}} \rangle_{\mathrm{HS}},
	\label{eq:decoupled_variance}
\end{align}
where the variance naturally decomposes as $V = V_1 + V_2$, where $V_1$ and $V_2$ denote the first-order projection component and the second-order residual component, respectively, given by
\begin{align}
	V_1 &:= \frac{4}{n}\langle \Delta\mu, C_{\mathbb{P}} \Delta\mu \rangle_{\mathcal{H}} + \frac{4}{m}\langle \Delta\mu, C_{\mathbb{Q}} \Delta\mu \rangle_{\mathcal{H}}, \label{eq:T1}\\
	V_2 &:= \frac{2}{n(n-1)}\|C_{\mathbb{P}}\|_{\mathrm{HS}}^2 + \frac{2}{m(m-1)}\|C_{\mathbb{Q}}\|_{\mathrm{HS}}^2 + \frac{4}{nm}\langle C_{\mathbb{P}}, C_{\mathbb{Q}} \rangle_{\mathrm{HS}}. \label{eq:T2}
\end{align}

Despite the closed-form expression in Eq.~\eqref{eq:decoupled_variance}, its finite-sample estimation is non-trivial. In particular, the direct plug-in estimator is biased, and constructing an unbiased estimator is challenging.

\subsection{Lossless acceleration for MMD with the Laplace kernel}
\label{sec:lossaccel}

Although MMD enjoys favorable theoretical properties, its naive computation relies on the explicit construction of kernel matrices, resulting in a computational complexity of $\mathcal{O}(d(n+m)^2)$. Letting $N=n+m$, this complexity scales as $\mathcal{O}(dN^2)$. For univariate samples, Bodenham and Kawahara \cite{bodenham2023eummd} proposed a lossless acceleration strategy, namely Efficient computation of univariate MMD (euMMD), which exploits the ordering structure of the data and reduces the complexity to $\mathcal{O}(N \log N)$.

The fast computation follows from permutation invariance of pairwise evaluations (Lemmas~\ref{lem1}--\ref{lem3}).

\begin{lemma}
	\label{lem1}
	Let $\{x_1,x_2,\dots,x_n\}\subset\mathbb{R}$ and let
	$f:\mathbb{R}\times\mathbb{R}\to\mathbb{R}$ be a symmetric function, i.e.,
	$f(x,x')=f(x',x)$. Then
	\begin{equation*}
		\sum_{i=1}^n \sum_{j=1}^i f(x_i,x_j) = \sum_{i=1}^n \sum_{j=i}^n f(x_i,x_j).
	\end{equation*}
\end{lemma}

\begin{lemma}
	\label{lem2}
	Under the conditions of Lemma 1, we have
	\begin{equation*}
		\sum_{i=1}^n \sum_{\substack{j=1\\ j\neq i}}^n f(x_i,x_j) = 2 \sum_{i=1}^n \sum_{j=1}^{i-1} f(x_i,x_j).
		\label{Applab1}
	\end{equation*}
\end{lemma}

\begin{lemma}
	\label{lem3}
	Let $\{x_1,x_2,\dots,x_n\}\subset\mathbb{R}$ and let
	$f:\mathbb{R}\times\mathbb{R}\to\mathbb{R}$ be a symmetric function, i.e.,
	$f(x,x')=f(x',x)$. Then for any permutation $\beta$ of $\{1,\dots,n\}$,
	we have
	\begin{equation*}
		\sum_{i=1}^n \sum_{j=1}^{i-1} f(x_i,x_j)
		=
		\sum_{i=1}^n \sum_{j=1}^{i-1} f\!\left(x_{\beta(i)},x_{\beta(j)}\right).
		\label{Applab2}
	\end{equation*}
\end{lemma}

With permutation invariance established by Lemmas~\ref{lem1}--\ref{lem3}, we assume without loss of generality that the samples are sorted. This allows the Laplacian kernel to be evaluated via a recursive decomposition, as stated in Proposition~\ref{pro4}.

\begin{proposition}
	\label{pro4}
	Let $\{x_1,\dots,x_n\}\subset\mathbb{R}$ be sorted such that $x_1 \le x_2 \le \dots \le x_n$.
	Consider the univariate Laplacian kernel with bandwidth $\sigma>0$,
	$k_{\mathrm{L}}(x, x'; \sigma) = \exp\!\left(-|x - x'|/\sigma\right)$.
	Define
	\begin{equation*}
		S := \sum_{i=1}^n \sum_{j=1}^{i-1} k_{\mathrm{L}}(x_i,x_j; \sigma)
		= \sum_{i=1}^n \sum_{j=1}^{i-1}
		\exp\!\left(-(x_i-x_j)/\sigma\right).
	\end{equation*}
	Then $S$ can be computed recursively as follows. Set $S_1=R_1=0$, and for
	$k=2,\dots,n$, define
	\begin{align*}
		D_k &= \exp\!\left(-(x_k-x_{k-1})/\sigma\right),\\
		R_k &= (R_{k-1}+1)\,D_k,\\
		S_k &= S_{k-1}+R_k.
	\end{align*}
	Then $S=S_n$. Sorting the sequence takes $\mathcal{O}(n \log n)$ time, and the subsequent recursive evaluation takes $\mathcal{O}(n)$ time, achieving an overall computational complexity of $\mathcal{O}(n \log n)$.
\end{proposition}

Lemmas~\ref{lem1}--\ref{lem2} reduce the quadratic forms in Eq.~\eqref{matrixMMD} to lower-triangular kernel sums:
\begin{align}
	\mathbf{1}_n^\top \tilde{\mathbf{K}}_{\mathbf{XX}} \mathbf{1}_n = 2T_1, \quad &\text{where} \quad T_1 = \sum_{i=2}^n \sum_{i'=1}^{i-1} k(x_i, x_{i'}), \\
	\mathbf{1}_m^\top \tilde{\mathbf{K}}_{\mathbf{YY}} \mathbf{1}_m = 2T_2, \quad &\text{where} \quad T_2 = \sum_{j=2}^m \sum_{j'=1}^{j-1} k(y_j, y_{j'}), \\
	\mathbf{1}_n^\top \mathbf{K}_{\mathbf{XY}} \mathbf{1}_m = T_3, \quad &\text{where} \quad T_3 = \sum_{i=1}^n \sum_{j=1}^m k(x_i, y_j).
\end{align}

For the Laplace kernel, $T_1$ and $T_2$ are computed via the recursive scheme in Proposition~\ref{pro4} after sorting $\mathbf{X}$ and $\mathbf{Y}$. For the cross term, let $\mathbf{Z}=\mathbf{X}\cup\mathbf{Y}$ with $N=n+m$. Applying the same recursion to $\mathbf{Z}$ yields
\begin{equation}
	T_4 = \sum_{k=2}^{n+m} \sum_{k'=1}^{k-1} k(z_k, z_{k'}) = \sum_{k=2}^{n+m} R_{k}^{\mathbf{Z}}.
\end{equation}

By construction, $T_4 = T_1 + T_2 + T_3$, hence
\begin{equation}
	T_3 = T_4 - T_1 - T_2.
\end{equation}

The overall complexity is $\mathcal{O}(N \log N)$ due to sorting, with $\mathcal{O}(N)$ memory usage.

\section{Unbiased estimation of the MMD variance and accelerated computation}
\label{sec3}
\subsection{Unbiased estimation of the MMD variance}
\label{sec31}

For notational convenience, we define $\nu_{\mathbb{P}} := \langle \Delta\mu, C_{\mathbb{P}} \Delta\mu \rangle_{\mathcal{H}}$ and $\nu_{\mathbb{Q}} := \langle \Delta\mu, C_{\mathbb{Q}} \Delta\mu \rangle_{\mathcal{H}}$ as the first-order variance components, and $\tau_{\mathbb{P}} := \|C_{\mathbb{P}}\|_{\mathrm{HS}}^2$, $\tau_{\mathbb{Q}} := \|C_{\mathbb{Q}}\|_{\mathrm{HS}}^2$, and $\tau_{\mathbb{PQ}} := \langle C_{\mathbb{P}}, C_{\mathbb{Q}} \rangle_{\mathrm{HS}}$ for the second-order Hilbert-Schmidt components.

Since the covariance operator $C_{\mathbb{P}}$ is self-adjoint, the cross-terms are identical, i.e., $\langle \mu_{\mathbb{Q}}, C_{\mathbb{P}} \mu_{\mathbb{P}} \rangle_{\mathcal{H}} = \langle \mu_{\mathbb{P}}, C_{\mathbb{P}} \mu_{\mathbb{Q}} \rangle_{\mathcal{H}}$, the expression simplifies to
\begin{equation}
	\nu_{\mathbb{P}} = \langle \mu_{\mathbb{P}}, C_{\mathbb{P}} \mu_{\mathbb{P}} \rangle_{\mathcal{H}} - 2 \langle \mu_{\mathbb{P}}, C_{\mathbb{P}} \mu_{\mathbb{Q}} \rangle_{\mathcal{H}} + \langle \mu_{\mathbb{Q}}, C_{\mathbb{P}} \mu_{\mathbb{Q}} \rangle_{\mathcal{H}}.
\end{equation}

By symmetry, $\nu_{\mathbb{Q}}$ can be analogously expanded as
\begin{equation}
	\nu_{\mathbb{Q}} = \langle \mu_{\mathbb{Q}}, C_{\mathbb{Q}} \mu_{\mathbb{Q}} \rangle_{\mathcal{H}} - 2 \langle \mu_{\mathbb{Q}}, C_{\mathbb{Q}} \mu_{\mathbb{P}} \rangle_{\mathcal{H}} + \langle \mu_{\mathbb{P}}, C_{\mathbb{Q}} \mu_{\mathbb{P}} \rangle_{\mathcal{H}}.
\end{equation}

For the second-order Hilbert--Schmidt terms $\tau_{\mathbb{P}}$, $\tau_{\mathbb{Q}}$, and $\tau_{\mathbb{PQ}}$, we use the covariance operator definition in Eq.~\ref{CP}, which can be equivalently rewritten as
\begin{equation}
	C_{\mathbb{P}} = \mathcal{M}_{\mathbb{P}} - \mu_{\mathbb{P}} \otimes \mu_{\mathbb{P}},
\end{equation}
where $\mathcal{M}_{\mathbb{P}} = \mathbb{E}_{X\sim\mathbb{P}} \big[ k(X,\cdot)\otimes k(X,\cdot) \big]$ denotes the uncentered second-moment operator.

Using the identity $\langle a\otimes b, c\otimes d\rangle_{\mathrm{HS}} = \langle a,c\rangle_{\mathcal{H}} \langle b,d\rangle_{\mathcal{H}}$, we expand
\begin{align}
	\tau_{\mathbb{P}} &= \|C_{\mathbb{P}}\|_{\mathrm{HS}}^2 = \left\| \mathcal{M}_{\mathbb{P}} - \mu_{\mathbb{P}}\otimes\mu_{\mathbb{P}} \right\|_{\mathrm{HS}}^2 \nonumber\\
	&= \|\mathcal{M}_{\mathbb{P}}\|_{\mathrm{HS}}^2 - 2 \left\langle \mathcal{M}_{\mathbb{P}}, \mu_{\mathbb{P}}\otimes\mu_{\mathbb{P}} \right\rangle_{\mathrm{HS}} + \| \mu_{\mathbb{P}}\otimes\mu_{\mathbb{P}} \|_{\mathrm{HS}}^2.
	\label{eq:tauP_expand}
\end{align}

The three terms on the right hand are computed as follows. First,
\begin{align}
	\|\mathcal{M}_{\mathbb{P}}\|_{\mathrm{HS}}^2 &= \left\langle \mathbb{E}_{X} [ \phi(X)\otimes\phi(X) ], \mathbb{E}_{X'} [ \phi(X')\otimes\phi(X') ] \right\rangle_{\mathrm{HS}} \nonumber\\
	&= \mathbb{E}_{X,X'} \left[ \langle \phi(X),\phi(X')\rangle_{\mathcal{H}}^2 \right] = \mathbb{E}_{X,X'} \left[ k(X,X')^2 \right].
	\label{eq:MHS}
\end{align}
where $\phi(X)=k(X,\cdot)$.

Next, using rank-one operator identity $\langle A, a \otimes b \rangle_{\mathrm{HS}} = \langle a, A b \rangle_{\mathcal{H}}$and the decomposition $\mathcal{M}_{\mathbb{P}} = C_{\mathbb{P}} + \mu_{\mathbb{P}}\otimes\mu_{\mathbb{P}}$, we obtain
\begin{align}
	\left\langle \mathcal{M}_{\mathbb{P}}, \mu_{\mathbb{P}}\otimes\mu_{\mathbb{P}} \right\rangle_{\mathrm{HS}} &= \langle \mu_{\mathbb{P}}, \mathcal{M}_{\mathbb{P}} \mu_{\mathbb{P}} \rangle_{\mathcal{H}} = \left\langle \mu_{\mathbb{P}}, (C_{\mathbb{P}} + \mu_{\mathbb{P}}\otimes\mu_{\mathbb{P}}) \mu_{\mathbb{P}} \right\rangle_{\mathcal{H}} \nonumber\\
	&= \langle \mu_{\mathbb{P}}, C_{\mathbb{P}}\mu_{\mathbb{P}} \rangle_{\mathcal{H}} + \langle \mu_{\mathbb{P}}, \mu_{\mathbb{P}} \rangle_{\mathcal{H}}^2.
	\label{eq:middle_decompose}
\end{align}

Then,
\begin{equation}
	\| \mu_{\mathbb{P}}\otimes\mu_{\mathbb{P}} \|_{\mathrm{HS}}^2 = \|\mu_{\mathbb{P}}\|_{\mathcal{H}}^4 = \langle \mu_{\mathbb{P}}, \mu_{\mathbb{P}} \rangle_{\mathcal{H}}^2.
	\label{eq:rankone}
\end{equation}

Substituting Eqs.~\eqref{eq:MHS}--\eqref{eq:rankone} into Eq.~\eqref{eq:tauP_expand}, we obtain
\begin{align}
	\tau_{\mathbb{P}} &= \mathbb{E}_{X,X'} [k(X,X')^2] - 2 \left( \langle \mu_{\mathbb{P}}, C_{\mathbb{P}}\mu_{\mathbb{P}} \rangle_{\mathcal{H}} + \langle \mu_{\mathbb{P}}, \mu_{\mathbb{P}} \rangle_{\mathcal{H}}^2 \right) + \langle \mu_{\mathbb{P}}, \mu_{\mathbb{P}} \rangle_{\mathcal{H}}^2 \nonumber\\
	&= \mathbb{E}_{X,X'} [k(X,X')^2] - 2 \langle \mu_{\mathbb{P}}, C_{\mathbb{P}}\mu_{\mathbb{P}} \rangle_{\mathcal{H}} - \langle \mu_{\mathbb{P}}, \mu_{\mathbb{P}} \rangle_{\mathcal{H}}^2.
\end{align}

By symmetry, the corresponding expression for $\mathbb{Q}$ is
\begin{equation}
	\tau_{\mathbb{Q}} = \mathbb{E}_{Y,Y'} [k(Y,Y')^2] - 2 \langle \mu_{\mathbb{Q}}, C_{\mathbb{Q}}\mu_{\mathbb{Q}} \rangle_{\mathcal{H}} - \langle \mu_{\mathbb{Q}}, \mu_{\mathbb{Q}} \rangle_{\mathcal{H}}^2.
\end{equation}

Finally, for the cross-covariance term $\tau_{\mathbb{PQ}} = \langle C_{\mathbb{P}}, C_{\mathbb{Q}} \rangle_{\mathrm{HS}}$, expanding $C_{\mathbb{P}} = \mathcal{M}_{\mathbb{P}} - \mu_{\mathbb{P}}\otimes\mu_{\mathbb{P}}$ and $C_{\mathbb{Q}} = \mathcal{M}_{\mathbb{Q}} - \mu_{\mathbb{Q}}\otimes\mu_{\mathbb{Q}}$ yields
\begin{equation}
	\tau_{\mathbb{PQ}} = \mathbb{E}_{X,Y}[k(X,Y)^2] - \langle \mu_{\mathbb{Q}}, C_{\mathbb{P}}\mu_{\mathbb{Q}} \rangle_{\mathcal{H}} - \langle \mu_{\mathbb{P}}, C_{\mathbb{Q}}\mu_{\mathbb{P}} \rangle_{\mathcal{H}} - \langle \mu_{\mathbb{P}}, \mu_{\mathbb{Q}} \rangle_{\mathcal{H}}^2.
\end{equation}

To estimate each variance component, we apply the inclusion-exclusion principle for generalized U-statistics \cite{sutherland2019unbiased, sane2013inclusion}. With the population-level expansions established, we construct empirical estimators by replacing kernel expectations and operator inner products with their corresponding matrix forms. As detailed in Appendix~\ref{App2}, this leads to explicit sub-estimators for the first-order terms:
\begin{align}
	\widehat{\nu}_{\mathbb{P}} &= \widehat{\langle \mu_{\mathbb{P}}, C_{\mathbb{P}} \mu_{\mathbb{P}} \rangle_{\mathcal{H}}} - 2 \widehat{\langle \mu_{\mathbb{P}}, C_{\mathbb{P}} \mu_{\mathbb{Q}} \rangle_{\mathcal{H}}} + \widehat{\langle \mu_{\mathbb{Q}}, C_{\mathbb{P}} \mu_{\mathbb{Q}} \rangle_{\mathcal{H}}} \nonumber\\
	&= \frac{1}{(n)_3} \left[ \|\tilde{\mathbf{K}}_{\mathbf{XX}} \mathbf{1}_n\|_2^2 - \|\tilde{\mathbf{K}}_{\mathbf{XX}}\|_F^2 \right] - \frac{2}{m (n)_2} \mathbf{1}_n^\top \tilde{\mathbf{K}}_{\mathbf{XX}} \mathbf{K}_{\mathbf{XY}} \mathbf{1}_m \nonumber \\
	&\quad + \frac{1}{n (m)_2} \left[ \|\mathbf{K}_{\mathbf{XY}} \mathbf{1}_m\|_2^2 - \|\mathbf{K}_{\mathbf{XY}}\|_F^2 \right] \nonumber \\
	&\quad - \frac{1}{(n)_4} \left[ (\mathbf{1}_n^\top \tilde{\mathbf{K}}_{\mathbf{XX}} \mathbf{1}_n)^2 - 4 \|\tilde{\mathbf{K}}_{\mathbf{XX}} \mathbf{1}_n\|_2^2 + 2 \|\tilde{\mathbf{K}}_{\mathbf{XX}}\|_F^2 \right] \nonumber \\
	&\quad + \frac{2}{m (n)_3} \left[ (\mathbf{1}_n^\top \tilde{\mathbf{K}}_{\mathbf{XX}} \mathbf{1}_n) (\mathbf{1}_n^\top \mathbf{K}_{\mathbf{XY}} \mathbf{1}_m) - 2 \mathbf{1}_n^\top \tilde{\mathbf{K}}_{\mathbf{XX}} \mathbf{K}_{\mathbf{XY}} \mathbf{1}_m \right] \nonumber \\
	&\quad - \frac{1}{(n)_2 (m)_2} \left[ (\mathbf{1}_n^\top \mathbf{K}_{\mathbf{XY}} \mathbf{1}_m)^2 - \|\mathbf{K}_{\mathbf{XY}}^\top \mathbf{1}_n\|_2^2 - \|\mathbf{K}_{\mathbf{XY}} \mathbf{1}_m\|_2^2 + \|\mathbf{K}_{\mathbf{XY}}\|_F^2 \right], \\
	\nonumber \\
	\widehat{\nu}_{\mathbb{Q}} &= \widehat{\langle \mu_{\mathbb{Q}}, C_{\mathbb{Q}} \mu_{\mathbb{Q}} \rangle_{\mathcal{H}}} - 2 \widehat{\langle \mu_{\mathbb{Q}}, C_{\mathbb{Q}} \mu_{\mathbb{P}} \rangle_{\mathcal{H}}} + \widehat{\langle \mu_{\mathbb{P}}, C_{\mathbb{Q}} \mu_{\mathbb{P}} \rangle_{\mathcal{H}}}\nonumber \\
	&= \frac{1}{(m)_3} \left[ \|\tilde{\mathbf{K}}_{\mathbf{YY}} \mathbf{1}_m\|_2^2 - \|\tilde{\mathbf{K}}_{\mathbf{YY}}\|_F^2 \right] - \frac{2}{n (m)_2} \mathbf{1}_m^\top \tilde{\mathbf{K}}_{\mathbf{YY}} \mathbf{K}_{\mathbf{XY}}^\top \mathbf{1}_n \nonumber \\
	&\quad + \frac{1}{m (n)_2} \left[ \|\mathbf{K}_{\mathbf{XY}}^\top \mathbf{1}_n\|_2^2 - \|\mathbf{K}_{\mathbf{XY}}\|_F^2 \right] \nonumber \\
	&\quad - \frac{1}{(m)_4} \left[ (\mathbf{1}_m^\top \tilde{\mathbf{K}}_{\mathbf{YY}} \mathbf{1}_m)^2 - 4 \|\tilde{\mathbf{K}}_{\mathbf{YY}} \mathbf{1}_m\|_2^2 + 2 \|\tilde{\mathbf{K}}_{\mathbf{YY}}\|_F^2 \right] \nonumber \\
	&\quad + \frac{2}{n (m)_3} \left[ (\mathbf{1}_m^\top \tilde{\mathbf{K}}_{\mathbf{YY}} \mathbf{1}_m) (\mathbf{1}_m^\top \mathbf{K}_{\mathbf{XY}}^\top \mathbf{1}_n) - 2 \mathbf{1}_m^\top \tilde{\mathbf{K}}_{\mathbf{YY}} \mathbf{K}_{\mathbf{XY}}^\top \mathbf{1}_n \right] \nonumber \\
	&\quad - \frac{1}{(n)_2 (m)_2} \left[ (\mathbf{1}_n^\top \mathbf{K}_{\mathbf{XY}} \mathbf{1}_m)^2 - \|\mathbf{K}_{\mathbf{XY}}^\top \mathbf{1}_n\|_2^2 - \|\mathbf{K}_{\mathbf{XY}} \mathbf{1}_m\|_2^2 + \|\mathbf{K}_{\mathbf{XY}}\|_F^2 \right].
\end{align}

Similarly, we construct unbiased estimators for the second-order Hilbert--Schmidt terms $\tau_{\mathbb{P}}$, $\tau_{\mathbb{Q}}$, and $\tau_{\mathbb{PQ}}$:
\begin{align}
	\widehat{\tau}_{\mathbb{P}} &= \widehat{\mathbb{E}_{X,X'} [k(X,X')^2]} - 2 \widehat{\langle \mu_{\mathbb{P}}, C_{\mathbb{P}} \mu_{\mathbb{P}} \rangle_{\mathcal{H}}} - \widehat{\langle \mu_{\mathbb{P}}, \mu_{\mathbb{P}} \rangle^2_{\mathcal{H}}} \nonumber \\
	&= \frac{1}{(n)_2} \|\tilde{\mathbf{K}}_{\mathbf{XX}}\|_F^2 - \frac{2}{(n)_3} \left[ \|\tilde{\mathbf{K}}_{\mathbf{XX}} \mathbf{1}_n\|_2^2 - \|\tilde{\mathbf{K}}_{\mathbf{XX}}\|_F^2 \right] \nonumber \\
	&\quad + \frac{1}{(n)_4} \left[ (\mathbf{1}_n^\top \tilde{\mathbf{K}}_{\mathbf{XX}} \mathbf{1}_n)^2 - 4 \|\tilde{\mathbf{K}}_{\mathbf{XX}} \mathbf{1}_n\|_2^2 + 2 \|\tilde{\mathbf{K}}_{\mathbf{XX}}\|_F^2 \right], \\
	\nonumber \\
	\widehat{\tau}_{\mathbb{Q}} &=  \widehat{\mathbb{E}_{Y,Y'} [k(Y,Y')^2]} - 2 \widehat{\langle \mu_{\mathbb{Q}}, C_{\mathbb{Q}} \mu_{\mathbb{Q}} \rangle_{\mathcal{H}}} - \widehat{\langle \mu_{\mathbb{Q}}, \mu_{\mathbb{Q}} \rangle^2_{\mathcal{H}}} \nonumber \\
	&= \frac{1}{(m)_2} \|\tilde{\mathbf{K}}_{\mathbf{YY}}\|_F^2 - \frac{2}{(m)_3} \left[ \|\tilde{\mathbf{K}}_{\mathbf{YY}} \mathbf{1}_m\|_2^2 - \|\tilde{\mathbf{K}}_{\mathbf{YY}}\|_F^2 \right] \nonumber \\
	&\quad + \frac{1}{(m)_4} \left[ (\mathbf{1}_m^\top \tilde{\mathbf{K}}_{\mathbf{YY}} \mathbf{1}_m)^2 - 4 \|\tilde{\mathbf{K}}_{\mathbf{YY}} \mathbf{1}_m\|_2^2 + 2 \|\tilde{\mathbf{K}}_{\mathbf{YY}}\|_F^2 \right], \\
	\nonumber \\
	\widehat{\tau}_{\mathbb{PQ}} &=  \widehat{\mathbb{E}_{X,Y}[k(X,Y)^2]}- \widehat{\langle \mu_{\mathbb{Q}}, C_{\mathbb{P}} \mu_{\mathbb{Q}} \rangle_{\mathcal{H}}} - \widehat{\langle \mu_{\mathbb{P}}, C_{\mathbb{Q}} \mu_{\mathbb{P}} \rangle_{\mathcal{H}}} - \widehat{\langle \mu_{\mathbb{P}}, \mu_{\mathbb{Q}} \rangle^2_{\mathcal{H}}} \nonumber \\
	&= \frac{1}{nm} \|\mathbf{K}_{\mathbf{XY}}\|_F^2 - \frac{1}{n (m)_2} \left[ \|\mathbf{K}_{\mathbf{XY}} \mathbf{1}_m\|_2^2 - \|\mathbf{K}_{\mathbf{XY}}\|_F^2 \right] \nonumber \\
	&\quad - \frac{1}{m (n)_2} \left[ \|\mathbf{K}_{\mathbf{XY}}^\top \mathbf{1}_n\|_2^2 - \|\mathbf{K}_{\mathbf{XY}}\|_F^2 \right] \nonumber \\
	&\quad + \frac{1}{(n)_2 (m)_2} \left[ (\mathbf{1}_n^\top \mathbf{K}_{\mathbf{XY}} \mathbf{1}_m)^2 - \|\mathbf{K}_{\mathbf{XY}}^\top \mathbf{1}_n\|_2^2 - \|\mathbf{K}_{\mathbf{XY}} \mathbf{1}_m\|_2^2 + \|\mathbf{K}_{\mathbf{XY}}\|_F^2 \right].
\end{align}

Finally, substituting these unbiased estimators into Eq.~\eqref{eq:decoupled_variance} yields the exact finite-sample variance estimator.

\begin{theorem}
	\label{thm:unbiased_variance}
	Let $\mathbf{X}=\{X_i\}_{i=1}^n$ and $\mathbf{Y}=\{Y_j\}_{j=1}^m$ be independent samples. The variance of the MMD statistic admits an exactly unbiased estimator of the form
	\begin{equation}
	\widehat{V} = \widehat{V}_1 + \widehat{V}_2,
    \end{equation}
	where
	\begin{align}
		\widehat{V}_1 &= \frac{4}{n}\widehat{\nu}_{\mathbb{P}} + \frac{4}{m}\widehat{\nu}_{\mathbb{Q}}, \\
		\widehat{V}_2 &= \frac{2}{(n)_2}\widehat{\tau}_{\mathbb{P}} + \frac{2}{(m)_2}\widehat{\tau}_{\mathbb{Q}} + \frac{4}{nm}\widehat{\tau}_{\mathbb{PQ}}.
	\end{align}
	Each term is constructed as a generalized U-statistic based unbiased estimator of its corresponding population quantity.
\end{theorem}

\begin{proof}
	Unbiasedness follows directly from the properties of generalized U-statistics.
	
	For any symmetric kernel $h$ on $k$ arguments, the associated U-statistic satisfies
	\begin{equation}
		\mathbb{E}\!\left[\frac{1}{(n)_k}\sum_{(i_1,\dots,i_k)} h(X_{i_1},\dots,X_{i_k})\right]
		= \mathbb{E}[h(X_1,\dots,X_k)].
	\end{equation}
	
	Each sub-estimator $\widehat{\nu}_{\mathbb{P}}, \widehat{\nu}_{\mathbb{Q}}, \widehat{\tau}_{\mathbb{P}}, \widehat{\tau}_{\mathbb{Q}}, \widehat{\tau}_{\mathbb{PQ}}$ is constructed as a symmetric kernel over distinct indices, with zero-diagonal Gram matrices ensuring exclusion of self-interactions. Hence, each estimator is unbiased for its corresponding population moment. By linearity of expectation, we obtain
	\begin{equation}
	\mathbb{E}[\widehat{V}]
	= \mathbb{E}[\widehat{V}_1] + \mathbb{E}[\widehat{V}_2]
	= V_1 + V_2
	= V.
    \end{equation}
\end{proof}

\subsection{Generalized quasi-linear acceleration method}

In \cite{bodenham2023eummd}, the symmetry of the kernel is exploited to construct a recursive prefix accumulator, which suffices for evaluating the unbiased MMD estimator in Eq.~\ref{matrixMMD}. However, computing the unbiased variance estimators in Theorem~\ref{thm:unbiased_variance} additionally requires explicit row sums of the within-sample Gram matrices (e.g., $\tilde{\mathbf{K}}_{\mathbf{XX}}\mathbf{1}_n$). To address this, we generalize the recursive scheme in Proposition~\ref{pro6} by maintaining both prefix ($R_i$) and suffix ($L_i$) accumulators.

\begin{proposition}
	\label{pro6}
	Let $\{x_1,\dots,x_n\}\subset\mathbb{R}$ be sorted such that $x_1 \le \cdots \le x_n$, and consider the Laplacian kernel $k_{\mathrm{L}}(x,x')=\exp(-|x-x'|/\sigma)$. Define $R_1=0$ and $L_n=0$. For $i=2,\dots,n$ and $i=n-1,\dots,1$, define the recursive accumulators
	\begin{align}
		D_i &= \exp\!\left(-(x_i-x_{i-1})/\sigma\right), \nonumber\\
		R_i &= (R_{i-1}+1)\,D_i, \\
		L_i &= (L_{i+1}+1)\,\exp\!\left(-(x_{i+1}-x_i)/\sigma\right). \nonumber
	\end{align}
	
	Then for each $i=1,\dots,n$, the off-diagonal row sum of the Laplace kernel matrix satisfies
	\begin{equation}
		\sum_{j\neq i} k_{\mathrm{L}}(x_i,x_j) = R_i + L_i.
	\end{equation}
\end{proposition}

\begin{proof}
	Since the sequence is sorted, the absolute value in the Laplace kernel reduces to directional differences. The off-diagonal row sum can be decomposed as
	\begin{equation}
	\sum_{j\neq i} k_{\mathrm{L}}(x_i,x_j)
	=
	\sum_{j<i} \exp\!\left(-(x_i-x_j)/\sigma\right)
	+
	\sum_{j>i} \exp\!\left(-(x_j-x_i)/\sigma\right).
    \end{equation}
	
	The first term is exactly the forward recursion $R_i$ by construction. The second term corresponds to the reverse-direction recursion, which follows the same update rule applied to the reversed sequence, yielding $L_i$. Summing both contributions gives the result.
\end{proof}

Algorithm~\ref{prefix_suffix} introduces explicit prefix and suffix accumulators for the Laplace kernel.

\begin{algorithm}
	\caption{Prefix-suffix accumulators for the Laplace kernel}
	\label{prefix_suffix}
	\begin{algorithmic}[1]
		\Require Ordered sample sequence $\mathbf{S} = \{s_1, \dots, s_n\}$, kernel parameter $\sigma$
		\Ensure Prefix accumulators $\{R_i\}_{i=1}^n$ and suffix accumulators $\{L_i\}_{i=1}^n$
		
		\State $R_1 \Leftarrow 0$; $L_n \Leftarrow 0$
		
		\For{$i = 2$ to $n$} 
		\State $D_i \Leftarrow \exp\!\bigl(-(s_i - s_{i-1})/\sigma\bigr)$
		\State $R_i \Leftarrow (R_{i-1} + 1) \cdot D_i$
		\EndFor
		
		\For{$i = n-1$ downto $1$}
		\State $D_{i+1} \Leftarrow \exp\!\bigl(-(s_{i+1} - s_{i})/\sigma\bigr)$
		\State $L_i \Leftarrow (L_{i+1} + 1) \cdot D_{i+1}$
		\EndFor
		
		\State \Return $\{R_i\}_{i=1}^n, \{L_i\}_{i=1}^n$
	\end{algorithmic}
\end{algorithm}

Extending this idea to cross-sample interactions, the previous implicit decomposition $T_3 = T_4 - T_1 - T_2$ is insufficient for variance acceleration, as it yields only a global scalar and does not retain the row-wise quantities required for constructing terms such as $\mathbf{K}_{\mathbf{XY}}\mathbf{1}_m$. To address this, we define prefix and suffix accumulators over the merged sequence, which explicitly preserve the cross-projection structure needed for unbiased variance computation.

\begin{proposition}
	\label{pro7}
	Let $x_1 \le \cdots \le x_n$ and $y_1 \le \cdots \le y_m$ be sorted samples, and let $\mathbf{Z} = \{z_k\}_{k=1}^{n+m}$ denote their merged sorted sequence with origin label $\ell(k)\in\{\mathbf{X},\mathbf{Y}\}$. Define forward and backward accumulators $\mathcal{A}_{x\to y}(k)$ and $\mathcal{Z}_{x\to y}(k)$ via
	\begin{align}
		\mathcal{A}_{x\to y}(k) &= \exp(-\Delta_k/\sigma)\mathcal{A}_{x\to y}(k-1) + \mathbb{I}\{\ell(k)=\mathbf{Y}\}, \\
		\mathcal{Z}_{x\to y}(k) &= \exp(-\Delta_{k+1}/\sigma)\mathcal{Z}_{x\to y}(k+1) + \mathbb{I}\{\ell(k)=\mathbf{Y}\},
	\end{align}
	where $\mathbb{I}\{\cdot\}$ denotes the indicator function. The boundary conditions are given by $\mathcal{A}_{x\to y}(1)=\mathbb{I}\{\ell(1)=\mathbf{Y}\}$ and $\mathcal{Z}_{x\to y}(n+m)=\mathbb{I}\{\ell(n+m)=\mathbf{Y}\}$.
	
	Then for any $x_i=z_k$, the cross-row sum satisfies
	\begin{equation}
	\sum_{j=1}^m k_{\mathrm{L}}(x_i,y_j) = \mathcal{A}_{x\to y}(k) + \mathcal{Z}_{x\to y}(k).
	\end{equation}
\end{proposition}

\begin{proof}
	For a target element $x_i \in \mathbf{X}$, let $k$ be its position in the merged sorted sequence $\mathbf{Z}$ (i.e., $z_k = x_i$). We consider the cross-row sum
	\begin{equation}
		S_i = \sum_{j=1}^m \exp\left(-|x_i - y_j|/\sigma\right).
	\end{equation}
		
	Since $\mathbf{Z}$ is sorted, the $\mathbf{Y}$-elements can be partitioned into those with indices $\le k$ and $> k$. Using the indicator $\mathbb{I}\{\ell(k')=\mathbf{Y}\}$, we obtain
	\begin{equation}
		\label{eq:split_sum}
		S_i =
		\sum_{k'=1}^k \mathbb{I}\{\ell(k')=\mathbf{Y}\}
		\exp\left(-(z_k - z_{k'})/\sigma\right)
		+
		\sum_{k'=k+1}^{n+m} \mathbb{I}\{\ell(k')=\mathbf{Y}\}
		\exp\left(-(z_{k'} - z_k)/\sigma\right).
	\end{equation}
	
	Define the prefix accumulator
	\begin{equation}
	\mathcal{A}_{x\to y}(k)=\sum_{k'=1}^k \mathbb{I}\{\ell(k')=\mathbf{Y}\}
	\exp\left(-(z_k - z_{k'})/\sigma\right),
    \end{equation}
    which satisfies the recursion
	\begin{align}
		\mathcal{A}_{x\to y}(k)
		&= \mathbb{I}\{\ell(k)=\mathbf{Y}\}
		+ \exp\left(-\Delta_k/\sigma\right)\mathcal{A}_{x\to y}(k-1),
	\end{align}
	with $\mathcal{A}_{x\to y}(1)=\mathbb{I}\{\ell(1)=\mathbf{Y}\}$.
		
	Similarly, define the suffix accumulator
	\begin{equation}
	\mathcal{Z}_{x\to y}(k)=\sum_{k'=k}^{n+m} \mathbb{I}\{\ell(k')=\mathbf{Y}\}
	\exp\left(-(z_{k'} - z_k)/\sigma\right),
	\end{equation}
	which satisfies
	\begin{align}
		\mathcal{Z}_{x\to y}(k)
		&= \mathbb{I}\{\ell(k)=\mathbf{Y}\}
		+ \exp\left(-\Delta_{k+1}/\sigma\right)\mathcal{Z}_{x\to y}(k+1),
	\end{align}
	with $\mathcal{Z}_{x\to y}(n+m)=\mathbb{I}\{\ell(n+m)=\mathbf{Y}\}$.
	
	Combining both parts in Eq.~\eqref{eq:split_sum}, we obtain
	\begin{equation}
		S_i = \mathcal{A}_{x\to y}(k) + \mathcal{Z}_{x\to y}(k)
		- \mathbb{I}\{\ell(k)=\mathbf{Y}\}.
	\end{equation}
	Since $x_i \in \mathbf{X}$ implies $\ell(k)=\mathbf{X}$, the indicator vanishes, yielding
	\begin{equation}
	S_i = \mathcal{A}_{x\to y}(k) + \mathcal{Z}_{x\to y}(k).
	\end{equation}
\end{proof}

The associated Algorithm~\ref{alg:cross_prefix_suffix} computes the cross-kernel prefix and suffix accumulators for both sequences in a single pass.

\begin{algorithm}[H]
	\caption{Cross-kernel prefix and suffix accumulators}
	\label{alg:cross_prefix_suffix}
	\begin{algorithmic}[1]
		\Require Sorted sequences $\mathbf{X} = \{x_1,\dots,x_n\}$, $\mathbf{Y} = \{y_1,\dots,y_m\}$; kernel parameter $\sigma$
		\Ensure Prefix and suffix accumulators: 
		$\mathcal{A}_{x\to y}, \mathcal{Z}_{x\to y}, \mathcal{A}_{y\to x}, \mathcal{Z}_{y\to x}$
		
		\State Merge $\mathbf{X}$ and $\mathbf{Y}$ into sorted sequence 
		$\mathbf{Z} = \{z_1,\dots,z_{n+m}\}$
		\State Record origin of each $z_k$ as $\ell(k) \in \{\mathbf{X}, \mathbf{Y}\}$
		\State Compute $\Delta_k = z_k - z_{k-1}$ for $k = 2,\dots,n+m$
		\State Precompute $\lambda_k = \exp(-\Delta_k/\sigma)$ for $k=2,\dots,n+m$
		
		\State $\mathcal{A}_{x\to y}[1] \Leftarrow 0$, \quad $\mathcal{A}_{y\to x}[1] \Leftarrow 0$
		\State $\mathcal{Z}_{x\to y}[n+m] \Leftarrow \mathbb{I}\{\ell(n+m)=\mathbf{Y}\}$
		\State $\mathcal{Z}_{y\to x}[n+m] \Leftarrow \mathbb{I}\{\ell(n+m)=\mathbf{X}\}$
		
		\For{$k = 2$ to $n+m$}
		\State $\mathcal{A}_{x\to y}[k] \Leftarrow 
		\lambda_k \cdot \mathcal{A}_{x\to y}[k-1] + \mathbb{I}\{\ell(k)=\mathbf{Y}\}$
		\State $\mathcal{A}_{y\to x}[k] \Leftarrow 
		\lambda_k \cdot \mathcal{A}_{y\to x}[k-1] + \mathbb{I}\{\ell(k)=\mathbf{X}\}$
		\EndFor
		
		\For{$k = n+m-1$ downto $1$}
		\State $\mathcal{Z}_{x\to y}[k] \Leftarrow 
		\lambda_{k+1} \cdot \mathcal{Z}_{x\to y}[k+1] + \mathbb{I}\{\ell(k)=\mathbf{Y}\}$
		\State $\mathcal{Z}_{y\to x}[k] \Leftarrow 
		\lambda_{k+1} \cdot \mathcal{Z}_{y\to x}[k+1] + \mathbb{I}\{\ell(k)=\mathbf{X}\}$
		\EndFor
		
		\State \Return $\mathcal{A}_{x\to y}, \mathcal{Z}_{x\to y}, \mathcal{A}_{y\to x}, \mathcal{Z}_{y\to x}$
	\end{algorithmic}
\end{algorithm}

Since the variance of MMD inherently requires computing squared kernel evaluations, such as the Frobenius norm $\|\mathbf{K}\|_F^2 = \sum k(x_i, x_j)^2$, the Lemma~\ref{lem8} and Proposition~\ref{pro9} are established.

\begin{lemma}
	\label{lem8}
	The $p$-th power of the Laplace kernel is equivalent to a Laplace kernel with bandwidth scaled by a factor of $1/p$. That is,
	\begin{equation}
		k_{\mathrm{L}}(x,x';\sigma)^p = k_{\mathrm{L}}(x,x';\sigma/p) = \exp\left( -p\frac{|x-x'|}{\sigma} \right).
	\end{equation}
\end{lemma}
\begin{proof}
	By the exponential power rule, $\left[\exp\left(-{|x-x'|}/{\sigma}\right)\right]^p = \exp\left(-{p|x-x'|}/{\sigma}\right)$.
\end{proof}

\begin{proposition}
	\label{pro9}
	Let $\{R_i^{\mathrm{sq}}, L_i^{\mathrm{sq}}\}_{i=1}^n$ and $\{\mathcal{A}_{x\to y}^{\mathrm{sq}}(k), \mathcal{Z}_{x\to y}^{\mathrm{sq}}(k)\}_{k=1}^{n+m}$ denote the accumulators from Propositions~\ref{pro6} and \ref{pro7}, evaluated with a halved bandwidth $\sigma/2$. If $k_i^x$ is the rank of $x_i$ in the merged sequence $\mathbf{Z}$, then
	\begin{align}
		\sum_{i=1}^n \sum_{j \neq i} k_{\mathrm{L}}(x_i, x_j; \sigma)^2 &= \sum_{i=1}^n \left( R_i^{\mathrm{sq}} + L_i^{\mathrm{sq}} \right), \\
		\sum_{i=1}^n \sum_{j=1}^m k_{\mathrm{L}}(x_i, y_j; \sigma)^2 &= \sum_{i=1}^n \left( \mathcal{A}_{x\to y}^{\mathrm{sq}}(k_i^x) + \mathcal{Z}_{x\to y}^{\mathrm{sq}}(k_i^x) \right).
	\end{align}
\end{proposition}

\begin{proof}
	By Lemma~\ref{lem8}, $k_{\mathrm{L}}(x, x'; \sigma)^2 = k_{\mathrm{L}}(x, x'; \sigma/2)$. Substituting this, the inner sum of the first equation reduces to $\sum_{j \neq i} k_{\mathrm{L}}(x_i, x_j; \sigma/2)$, which equals $R_i^{\mathrm{sq}} + L_i^{\mathrm{sq}}$ by Proposition~\ref{pro6}. Summing over $i$ yields the first identity. The second identity follows identically by applying Proposition~\ref{pro7} to the cross-sample inner sum.
\end{proof}

The computational complexity of this generalized procedure is dominated by sorting $\mathbf{X}$ and $\mathbf{Y}$, requiring $\mathcal{O}(n \log n + m \log m)$ operations, which can be equivalently expressed as $\mathcal{O}(N \log N)$ with $N=n+m$. Once sorted, all subsequent steps, including sequence merging and computation of prefix/suffix accumulators, are performed in linear time $\mathcal{O}(N)$ via sequential scans. Importantly, the extraction of localized row sums required for exact variance estimation introduces no additional asymptotic overhead, preserving the overall $\mathcal{O}(N \log N)$ complexity of the original euMMD framework.

\subsection{Accelerated computation of MMD variance estimator}
\label{sec33}
Let $x_1 \le \dots \le x_n$ and $y_1 \le \dots \le y_m$ be the sorted samples. According to the Proposition~\ref{pro6},~\ref{pro7}, and ~\ref{pro9}, define the base and squared prefix/suffix row sums for the single-sample kernel matrices: 
$$ R_i^{\mathbf X}, L_i^{\mathbf X}, R_i^{\mathrm{sq}, \mathbf X}, L_i^{\mathrm{sq}, \mathbf X}, \quad i=1,\dots,n; \qquad R_j^{\mathbf Y}, L_j^{\mathbf Y}, R_j^{\mathrm{sq}, \mathbf Y}, L_j^{\mathrm{sq}, \mathbf Y}, \quad j=1,\dots,m, $$ 
and the corresponding base and squared cross-sample accumulators over the merged sequence: 
$$ \mathcal{A}_{x\to y}(k), \mathcal{Z}_{x\to y}(k), \mathcal{A}^{\mathrm{sq}}_{x\to y}(k), \mathcal{Z}^{\mathrm{sq}}_{x\to y}(k) \quad \text{and} \quad \mathcal{A}_{y\to x}(k), \mathcal{Z}_{y\to x}(k), \mathcal{A}^{\mathrm{sq}}_{y\to x}(k), \mathcal{Z}^{\mathrm{sq}}_{y\to x}(k), $$
where $k=1,\dots,n+m$.

The necessary matrix traces, Frobenius norms, and cross-projection terms required by the unbiased variance sub-estimators derived in Section~\ref{sec31} can be explicitly computed as simple linear combinations and element-wise products of the accumulators defined above.

\paragraph{Single-sample kernel matrix}
For the $\mathbf{X}$ single-sample kernel matrix:
\begin{align}
	\|\tilde{\mathbf{K}}_{\mathbf{XX}}\|_F^2 &= \sum_{i=1}^n \big(R_i^{\mathrm{sq}, \mathbf X} + L_i^{\mathrm{sq}, \mathbf X}\big),\\
	\|\tilde{\mathbf{K}}_{\mathbf{XX}}\mathbf{1}_n\|_2^2 &= \sum_{i=1}^n \big(R_i^{\mathbf X} + L_i^{\mathbf X}\big)^2,\\
	(\mathbf{1}_n^\top \tilde{\mathbf{K}}_{\mathbf{XX}}\mathbf{1}_n)^2 &= \Big(\sum_{i=1}^n \big(R_i^{\mathbf X} + L_i^{\mathbf X}\big)\Big)^2.
\end{align}

For the $\mathbf{Y}$ single-sample kernel matrix:
\begin{align}
	\|\tilde{\mathbf{K}}_{\mathbf{YY}}\|_F^2 &= \sum_{j=1}^m \big(R_j^{\mathrm{sq}, \mathbf Y} + L_j^{\mathrm{sq}, \mathbf Y}\big),\\
	\|\tilde{\mathbf{K}}_{\mathbf{YY}}\mathbf{1}_m\|_2^2 &= \sum_{j=1}^m \big(R_j^{\mathbf Y} + L_j^{\mathbf Y}\big)^2,\\
	(\mathbf{1}_m^\top \tilde{\mathbf{K}}_{\mathbf{YY}}\mathbf{1}_m)^2 &= \Big(\sum_{j=1}^m \big(R_j^{\mathbf Y} + L_j^{\mathbf Y}\big)\Big)^2.
\end{align}

\paragraph{Cross-sample kernel matrix}
For the cross-sample accumulators, let $k_i^x$ and $k_j^y$ denote the indices of $x_i$ and $y_j$ in the merged sorted sequence $\mathbf{Z}$, respectively. For notational convenience, we define the shorthand $\mathcal{A}_{x\to y}(i) := \mathcal{A}_{x\to y}(k_i^x)$, and apply the same mapping to the other cross-accumulators. Then, for the cross-sample kernel matrix:
\begin{align}
	\|\mathbf{K}_{\mathbf{XY}}\|_F^2 &= \sum_{i=1}^n \big(\mathcal{A}^{\mathrm{sq}}_{x\to y}(i) + \mathcal{Z}^{\mathrm{sq}}_{x\to y}(i)\big),\\
	\|\mathbf{K}_{\mathbf{XY}}\mathbf{1}_m\|_2^2 &= \sum_{i=1}^n \big(\mathcal{A}_{x\to y}(i) + \mathcal{Z}_{x\to y}(i)\big)^2, \\
	\|\mathbf{K}_{\mathbf{XY}}^\top \mathbf{1}_n\|_2^2 &= \sum_{j=1}^m \big(\mathcal{A}_{y\to x}(j) + \mathcal{Z}_{y\to x}(j)\big)^2,\\
	(\mathbf{1}_n^\top \mathbf{K}_{\mathbf{XY}} \mathbf{1}_m)^2 &= \Big(\sum_{i=1}^n \big(\mathcal{A}_{x\to y}(i) + \mathcal{Z}_{x\to y}(i)\big)\Big)^2.
\end{align}

\paragraph{Mixed cross-covariance projections}
Crucially, evaluating the first-order variance components requires the mixed cross-covariance projections, these terms are efficiently obtained as:
\begin{align}
	\mathbf{1}_n^\top \tilde{\mathbf{K}}_{\mathbf{XX}} \mathbf{K}_{\mathbf{XY}} \mathbf{1}_m &= \sum_{i=1}^n \big(R_i^{\mathbf X} + L_i^{\mathbf X}\big) \cdot \big(\mathcal{A}_{x\to y}(i) + \mathcal{Z}_{x\to y}(i)\big), \\
	\mathbf{1}_m^\top \tilde{\mathbf{K}}_{\mathbf{YY}} \mathbf{K}_{\mathbf{XY}}^\top \mathbf{1}_n &= \sum_{j=1}^m \big(R_j^{\mathbf Y} + L_j^{\mathbf Y}\big) \cdot \big(\mathcal{A}_{y\to x}(j) + \mathcal{Z}_{y\to x}(j)\big).
\end{align}

Based on this theoretical breakdown, Algorithm~\ref{alg:fast_mmd} presents the complete accelerated procedure for computing the unbiased MMD along with its exact variance estimator, seamlessly reducing the overall computational complexity from the naive $\mathcal{O}(N^2)$ to $\mathcal{O}(N \log N)$.

\begin{algorithm}[H]
	\caption{Extended euMMD for unbiased MMD$^2$ and exact variance estimation}
	\label{alg:fast_mmd}
	\begin{algorithmic}[1]
		\Require Samples $\mathbf{X}$, $\mathbf{Y}$ of sizes $n, m$; kernel parameter $\sigma$
		\Ensure Unbiased $\widehat{\mathrm{MMD}}^2$ and its variance estimate $\widehat{V} = \widehat{V}_1 + \widehat{V}_2$
		
		\Statex $\diamond$ \textbf{Base Accumulators ($\sigma$)}
		\State Sort $\mathbf{X}, \mathbf{Y} \to \mathbf{X}_s, \mathbf{Y}_s$
		\State $(R^{\mathbf X},L^{\mathbf X}) \Leftarrow \textsc{PrefixSuffix}(\mathbf{X}_s,\sigma)$
		\State $(R^{\mathbf Y},L^{\mathbf Y}) \Leftarrow \textsc{PrefixSuffix}(\mathbf{Y}_s,\sigma)$
		\State $(\mathcal{A}_{x\to y}, \mathcal{Z}_{x\to y}, \dots) \Leftarrow \textsc{CrossPrefixSuffix}(\mathbf{X}_s,\mathbf{Y}_s,\sigma)$
		
		\Statex $\diamond$ \textbf{Squared Accumulators ($\sigma/2$)}
		\State $(R^{\text{sq}, \mathbf X},L^{\text{sq}, \mathbf X}) \Leftarrow \textsc{PrefixSuffix}(\mathbf{X}_s,\sigma/2)$
		\State $(R^{\text{sq}, \mathbf Y},L^{\text{sq}, \mathbf Y}) \Leftarrow \textsc{PrefixSuffix}(\mathbf{Y}_s,\sigma/2)$
		\State $(\mathcal{A}^{\text{sq}}_{x\to y}, \mathcal{Z}^{\text{sq}}_{x\to y}, \dots) \Leftarrow \textsc{CrossPrefixSuffix}(\mathbf{X}_s,\mathbf{Y}_s,\sigma/2)$
		
		\Statex $\diamond$ \textbf{Final MMD \& Variance Assembly}
		\State Compute $T_1$, $T_2$, and $T_3$ using linear sums of base accumulators
		\State $\widehat{\mathrm{MMD}}^2 \Leftarrow \frac{2T_1}{(n)_2} + \frac{2T_2}{(m)_2} - \frac{2T_3}{nm}$
		\State Extract matrix norms and mixed projections via linear sums of accumulators
		\State Compute $\widehat{\nu}_{\mathbb{P}}, \widehat{\nu}_{\mathbb{Q}}, \widehat{\tau}_{\mathbb{P}}, \widehat{\tau}_{\mathbb{Q}}, \widehat{\tau}_{\mathbb{PQ}}$ using the extracted terms in Sections~\ref{sec31} and~\ref{sec33}.
		\State $\widehat{V}_1 \Leftarrow \frac{4}{n}\widehat{\nu}_{\mathbb{P}} + \frac{4}{m}\widehat{\nu}_{\mathbb{Q}}$
		\State $\widehat{V}_2 \Leftarrow \frac{2}{(n)_2}\widehat{\tau}_{\mathbb{P}} + \frac{2}{(m)_2}\widehat{\tau}_{\mathbb{Q}} + \frac{4}{nm}\widehat{\tau}_{\mathbb{PQ}}$
		
		\State \Return $\widehat{\mathrm{MMD}}^2$, $\widehat{V}_1 + \widehat{V}_2$
	\end{algorithmic}
\end{algorithm}

\section{Numerical experiments}
\label{sec4}
\subsection{Variance analysis under heterogeneous settings}

\begin{figure}[htb]
	\centering
	\includegraphics{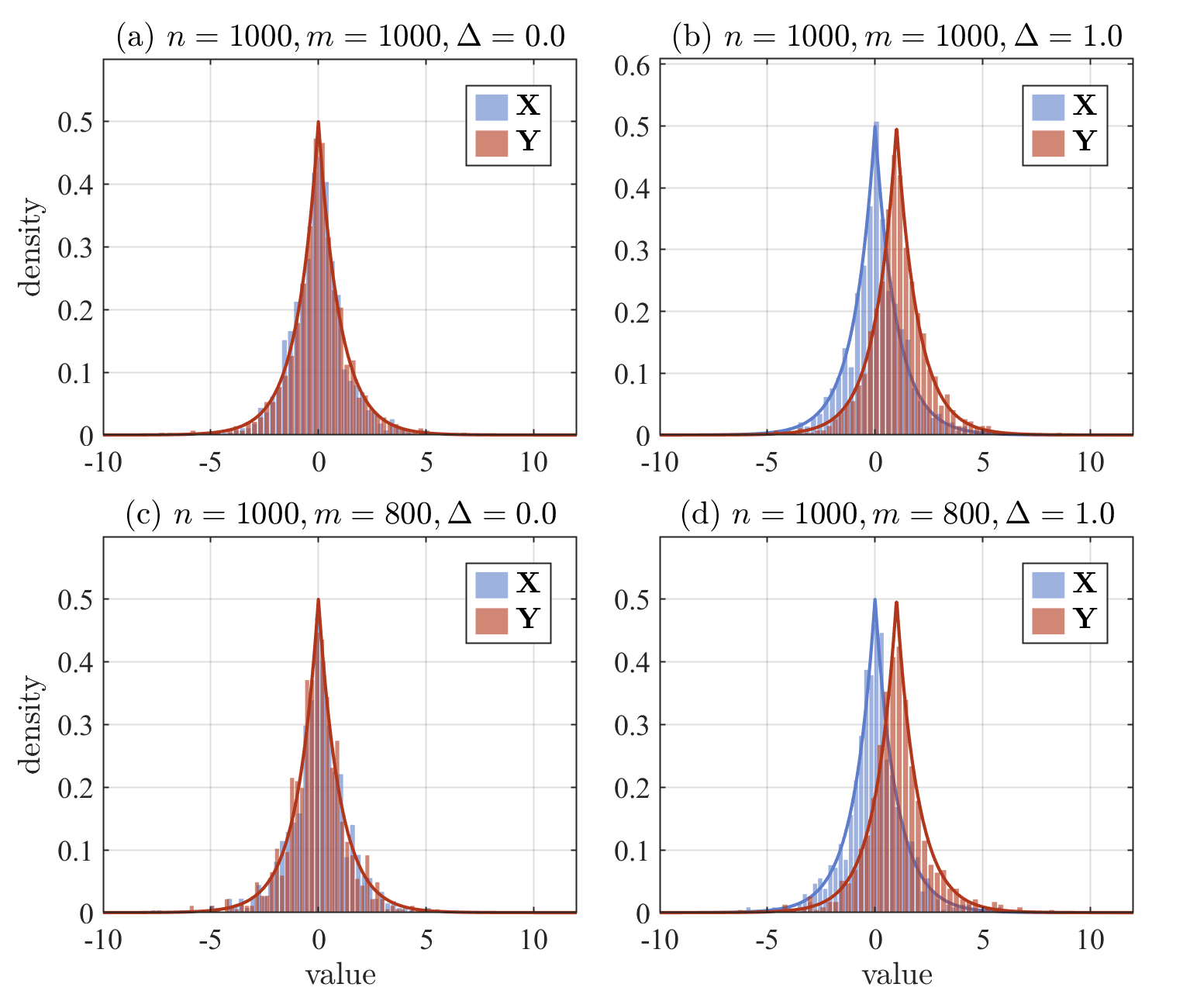}
	\caption{Empirical sample distributions in different cases. The blue histograms and curves represent the base sample $\mathbf{X} \sim \mathbb{P}$, while the red ones correspond to the shifted sample $\mathbf{Y} \sim \mathbb{Q}$.}
	\label{fig:sample_dist}
\end{figure}

To evaluate the proposed MMD variance estimator, we conduct benchmarks using samples drawn from $\mathbf{X} \sim \mathbb{P} = \text{Laplace}(0, 1)$ and $\mathbf{Y} \sim \mathbb{Q} = \text{Laplace}(\Delta, 1)$. The sample size of $\mathbf{X}$ is fixed at $n$. We manipulate the sample size $m$ of $\mathbf{Y}$ and the location shift $\Delta$ to construct four distinct settings across sample scales $n \in \{316, 1000, 3162, 10000\}$: 
Case A ($m = n, \Delta = 0$), Case B ($m = n, \Delta = 1$), Case C ($m = 0.8n, \Delta = 0$), and Case D ($m = 0.8n, \Delta = 1$). Figure~\ref{fig:sample_dist} illustrates their empirical distributions.

\paragraph{Case A: Null hypothesis with balanced samples ($m=n, \Delta = 0$)}
We benchmark against the classical estimator of Gretton et al.~\cite{gretton2012kernel} in Table~\ref{tab:caseA}. Under the degenerate null hypothesis, Gretton's asymptotic formulation theoretically eliminates the vanishing first-order contributions and evaluates only the second-order interactions. Correspondingly, our explicitly second-order variance exhibits near-identical agreement with Gretton's baseline across all sample scales.

\begin{table}[h]
	\centering
	\caption{Comparison of MMD and variance estimators under equal sample sizes ($m=n$) under the null hypothesis $\mathcal{H}_0$}
	\label{tab:caseA}
	\begin{tabular}{llcccc}
		\toprule
		Kernel & Estimator & $n=316$ & $n=1000$ & $n=3162$ & $n=10000$ \\
		\midrule
		\textit{Laplacian} 
		& MMD Mean    & 1.543e-04 & -5.979e-04 & 7.656e-05 & -6.029e-05 \\
		& Gretton et al.          & 3.520e-06 & 3.864e-07  & 3.955e-08 & 3.807e-09  \\
		& Proposed (Second-order) & 3.900e-06 & 3.819e-07  & 3.881e-08 & 3.859e-09  \\
		& Proposed (Total)        & 5.539e-06 & -2.066e-07 & 5.281e-08 & -2.044e-09 \\
		\midrule
		\textit{Gaussian} 
		& MMD Mean    & 6.421e-04 & -5.143e-04 & 2.011e-04 & -6.921e-05 \\
		& Gretton et al.          & 5.209e-06 & 5.797e-07  & 6.041e-08 & 5.725e-09  \\
		& Proposed (Second-order) & 5.888e-06 & 5.739e-07  & 5.893e-08 & 5.839e-09  \\
		& Proposed (Total)        & 8.976e-06 & -2.716e-07 & 1.309e-07 & -4.178e-09 \\
		\bottomrule
	\end{tabular}
\end{table}

\paragraph{Case B: Alternative hypothesis with balanced samples ($m = n, \Delta \neq 0$)}
We evaluate the precision under $\mathcal{H}_1$ against the exact classical unbiased estimator of Sutherland et al.~\cite{sutherland2019unbiased} in Table~\ref{tab:caseC}. The absolute variance discrepancy remains entirely negligible (e.g., on the order of $10^{-13}$ to $10^{-9}$), confirming identical dominant variance under $\mathcal{H}_1$. This agreement originates from the fact that Sutherland et al. employ an uncentered covariance operator, whose effect is reflected in the slightly different term-wise decomposition presented in Section~\ref{B2}.

\begin{table}[h]
	\centering
	\caption{Comparison of MMD and variance estimators under equal sample sizes ($m=n$) under the alternative hypothesis $\mathcal{H}_1$ ($\Delta = 1$)}
	\label{tab:caseC}
	\begin{tabular}{llcccc}
		\toprule
		Kernel & Estimator & $n=316$ & $n=1000$ & $n=3162$ & $n=10000$ \\
		\midrule
		\textit{Laplacian} 
		& MMD Mean                & 9.584e-02 & 9.906e-02 & 1.017e-01 & 1.001e-01 \\ % Sample value for n=10000
		& Sutherland et al.       & 3.473e-04 & 1.093e-04 & 3.578e-05 & 1.131e-05 \\
		& Proposed        & 3.473e-04 & 1.093e-04 & 3.578e-05 & 1.131e-05 \\
		& Abs difference       & 5.309e-09 & 1.603e-10 & 5.142e-12 & 1.542e-13 \\
		\midrule
		\textit{Gaussian} 
		& MMD Mean                & 1.286e-01 & 1.281e-01 & 1.344e-01 & 1.312e-01 \\
		& Sutherland et al.       & 6.408e-04 & 1.930e-04 & 6.465e-05 & 2.045e-05 \\
		& Proposed        & 6.408e-04 & 1.930e-04 & 6.465e-05 & 2.045e-05 \\
		& Abs difference       & 8.621e-09 & 2.532e-10 & 8.229e-12 & 2.451e-13 \\
		\bottomrule
	\end{tabular}
\end{table}

\paragraph{Case C: Null hypothesis with unbalanced samples ($m\neq n, \Delta = 0$)}
Variance estimation under asymmetric regimes ($m \neq n$) remains largely unexplored. While Gretton's estimator is restricted to the balanced setting $m=n$, the recent extension of Wei et al.~\cite{wei2025maximum} adopts a biased formulation that retains only the first-order component. Since this component theoretically vanishes under $\mathcal{H}_0$, existing approaches become unstable near the degenerate boundary and may produce negative variance estimates. Table~\ref{tab:caseB} evaluates our method under a fixed ratio $m=0.8n$, the proposed second-order estimator remains strictly nonnegative across all scales, demonstrating robust stability under degeneracy.

\begin{table}[h]
	\centering
	\caption{Proposed MMD variance estimators under unbalanced sample ratio ($m = 0.8n$) under the null hypothesis $\mathcal{H}_0$}
	\label{tab:caseB}
	\begin{tabular}{llcccc}
		\toprule
		Kernel & Estimator & $n=316$ & $n=1000$ & $n=3162$ & $n=10000$ \\
		\midrule
		\textit{Laplacian} 
		& MMD Mean    & 1.097e-03 & 4.589e-04 & 3.927e-04 & 4.153e-05 \\
		& Proposed (Second-order) & 4.918e-06 & 4.814e-07 & 4.893e-08 & 4.853e-09 \\
		& Proposed (Total)        & 2.268e-06 & 1.103e-06 & 2.116e-07 & 1.164e-08 \\
		\midrule
		\textit{Gaussian} 
		& MMD Mean    & -8.667e-04 & 1.857e-04 & 3.918e-04 & 5.696e-05 \\
		& Proposed (Second-order) & 7.431e-06  & 7.264e-07 & 7.420e-08 & 7.322e-09 \\
		& Proposed (Total)        & -4.209e-06 & 1.196e-06 & 2.514e-07 & 2.200e-08 \\
		\bottomrule
	\end{tabular}
\end{table}

\paragraph{Case D: Alternative hypothesis with unbalanced samples ($m\neq n, \Delta \neq 0$)}
In this general non-degenerate regime, the first-order linear projection heavily dominates the variance profile, and only the asymptotic framework of Wei et al. \cite{wei2025maximum} and our formulation are applicable. Setting $m = 0.8n$, Table~\ref{tab:caseD_1} demonstrates an exceptional numerical parity between our first-order component and Wei et al.'s baseline. Crucially, their methodology inherently remains a biased approximation, whereas our proposed framework delivers a strictly unbiased finite-sample estimator.

\begin{table}[h]
	\centering
	\caption{Comparison of MMD variance estimators under unbalanced alternative hypothesis ($\Delta = 1, m = 0.8n$)}
	\label{tab:caseD_1}
	\begin{tabular}{llcccc}
		\toprule
		Kernel & Estimator & $n=316$ & $n=1000$ & $n=3162$ & $n=10000$ \\
		\midrule
		\textit{Laplacian} 
		& MMD Mean    & 1.243e-01 & 9.634e-02 & 9.895e-02 & 1.044e-01 \\
		& Wei et al.              & 4.818e-04 & 1.275e-04 & 4.000e-05 & 1.316e-05 \\
		& Proposed (First-order)  & 4.782e-04 & 1.270e-04 & 3.995e-05 & 1.315e-05 \\
		& Proposed (Second-order) & 4.306e-06 & 4.583e-07 & 4.523e-08 & 4.487e-09 \\
		\midrule
		\textit{Gaussian} 
		& MMD Mean   & 1.636e-01 & 1.297e-01 & 1.303e-01 & 1.390e-01 \\
		& Wei et al.              & 8.733e-04 & 2.343e-04 & 7.154e-05 & 2.396e-05 \\
		& Proposed (First-order)  & 8.691e-04 & 2.336e-04 & 7.148e-05 & 2.396e-05 \\
		& Proposed (Second-order) & 6.492e-06 & 7.011e-07 & 6.995e-08 & 6.930e-09 \\
		\bottomrule
	\end{tabular}
\end{table}

\subsection{Validation of the estimator properties}
Building on the theoretical framework of \cite{wei2025maximum} for MMD degeneracy and asymptotic behavior, this section connects the theory with finite-sample practice. We restate the main results and validate them through simulations, showing that the predicted degeneracy and convergence behaviors are well reflected empirically.

\paragraph{Variance decomposition and structural transition}
\begin{theorem}
	\label{prop:degen-conds}
	Let $\widehat{\mathrm{MMD}}^2$ be the MMD estimator for distributions $\mathbb{P}$ and $\mathbb{Q}$. Assuming at least one of the covariance operators $C_{\mathbb{P}}$ or $C_{\mathbb{Q}}$ is non-zero, the order of degeneracy is determined as follows:
	\begin{enumerate}
		\item If $\mu_{\mathbb{P}} = \mu_{\mathbb{Q}}$, $\widehat{\mathrm{MMD}}^2$ is first-order degenerate,
		
		\item If $\mu_{\mathbb{P}} \ne \mu_{\mathbb{Q}}$, $\widehat{\mathrm{MMD}}^2$ is typically non-degenerate (variance decays as $\mathcal{O}(n^{-1})$).
	\end{enumerate}
\end{theorem}

\begin{figure}[htbp]
	\centering
	\includegraphics{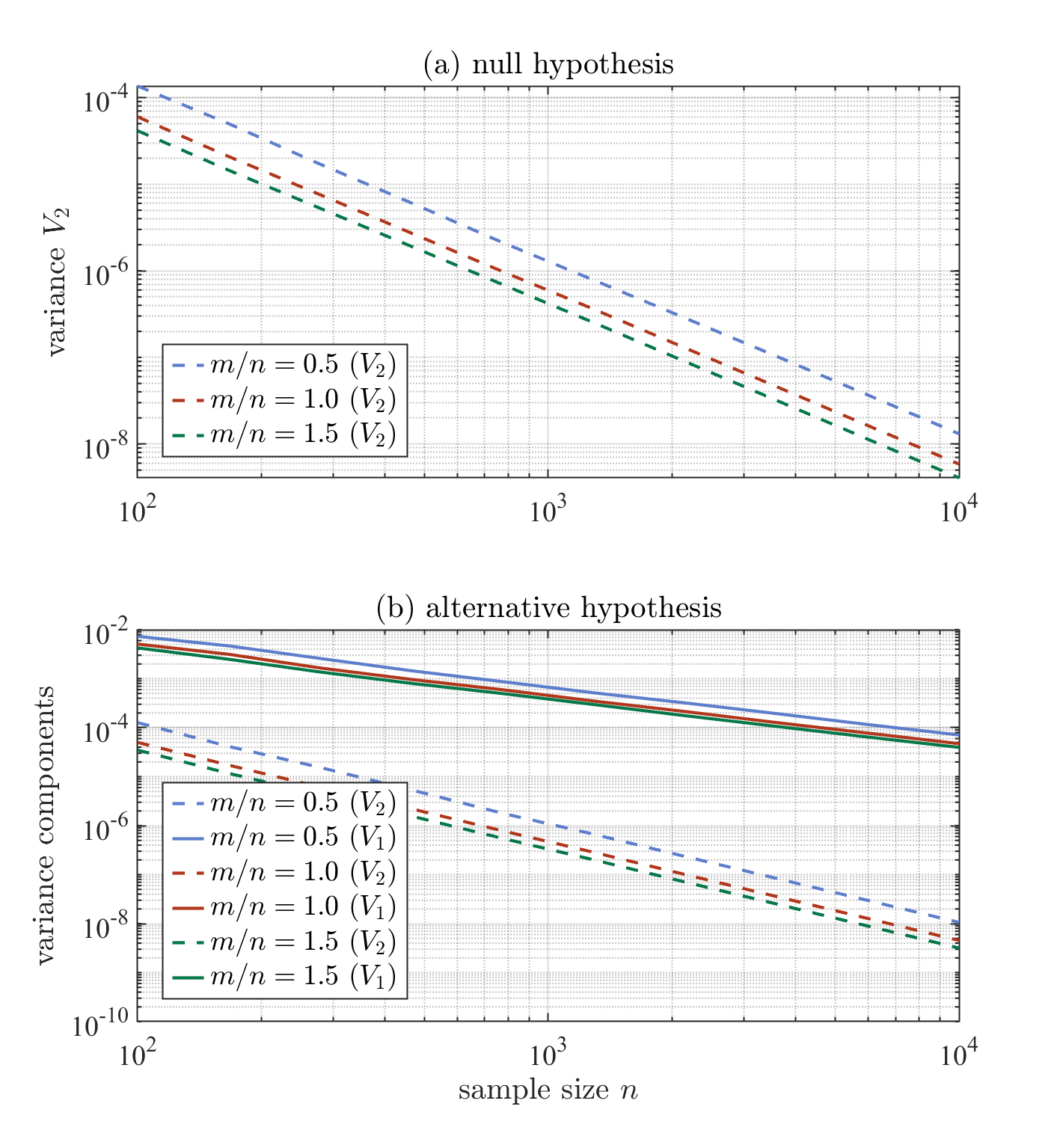}
	\caption{Variance decomposition of the MMD estimator. (a) $V_2$ component decay under Null hypotheses. (b) Dominance of $V_1$ over $V_2$ under the Alternative hypothesis.}
	\label{fig:variance_combined}
\end{figure}

Figure~\ref{fig:variance_combined} empirically validates Theorem~\ref{prop:degen-conds} by confirming the distinct variance decay rates under different hypotheses. In our simulation setup, we draw samples from one-dimensional Gaussian distributions: $\mathbf{X}, \mathbf{Y} \sim \mathcal{N}(0, 1)$ for the null regime, and $\mathbf{Y} \sim \mathcal{N}(1.5, 1)$ for the alternative regime. We employ a Gaussian kernel with bandwidth $\sigma = 1.0$ and evaluate the variance across sample sizes $n$ ranging from $10^2$ to $10^4$, with varying sample size ratios $m/n \in \{0.5, 1.0, 1.5\}$. Specifically, the simulations confirm that first-order degeneracy leads to an $\mathcal{O}(n^{-2})$ variance decay, while the non-degenerate regime exhibits the standard $\mathcal{O}(n^{-1})$ scaling. Although the second-order contribution retains an $\mathcal{O}(n^{-2})$ scaling, it is asymptotically dominated by the first-order term and therefore does not affect the leading-order variance behavior.

\paragraph{Asymptotic normality under unbalanced sampling}
\begin{corollary}
	\label{thm:alt-dist}
	Assume the alternative hypothesis holds, i.e., $\mathbb{P} \neq \mathbb{Q}$. Let the sample sizes grow such that $\min\{n, m\}/n \to \rho_X$ and $\min\{n, m\}/m \to \rho_Y$ for some constants $\rho_X, \rho_Y \in [0, 1]$. Then, the estimator $\widehat{\mathrm{MMD}}^2$ is asymptotically normal, satisfying:
	\begin{equation}
		\sqrt{\min\{n, m\}}(\widehat{\mathrm{MMD}}^2 - \mathrm{MMD}^2(\mathbb{P}, \mathbb{Q})) \xrightarrow{d} \mathcal{N}(0, 4\rho_X \nu_{\mathbb{P}} + 4 \rho_Y \nu_{\mathbb{Q}}),
	\end{equation}
	where $\nu_{\mathbb{P}} = \langle \Delta\mu, C_{\mathbb{P}} \Delta\mu \rangle_{\mathcal{H}}$ and $\nu_{\mathbb{Q}} = \langle \Delta\mu, C_{\mathbb{Q}} \Delta\mu \rangle_{\mathcal{H}}$.
\end{corollary}

To empirically validate Corollary~\ref{thm:alt-dist}, we conduct a large-scale Monte Carlo simulation consisting of $K=5,000$ independent trials. We fix the base sample size at $n=1,000$ and vary the secondary sample size $m \in \{500, 700, 2,000, 3,000\}$ to evaluate the estimator across four distinct sample size ratios ($m/n \in \{0.5, 0.7, 2.0, 3.0\}$). Following the alternative hypothesis setup, the underlying distributions are specified as $\mathbf{X} \sim \mathcal{N}(0, 1)$ and $\mathbf{Y} \sim \mathcal{N}(1.5, 1)$, utilizing a Gaussian kernel with bandwidth $\sigma=1.0$.

\begin{figure}[htbp]
	\centering
	\includegraphics{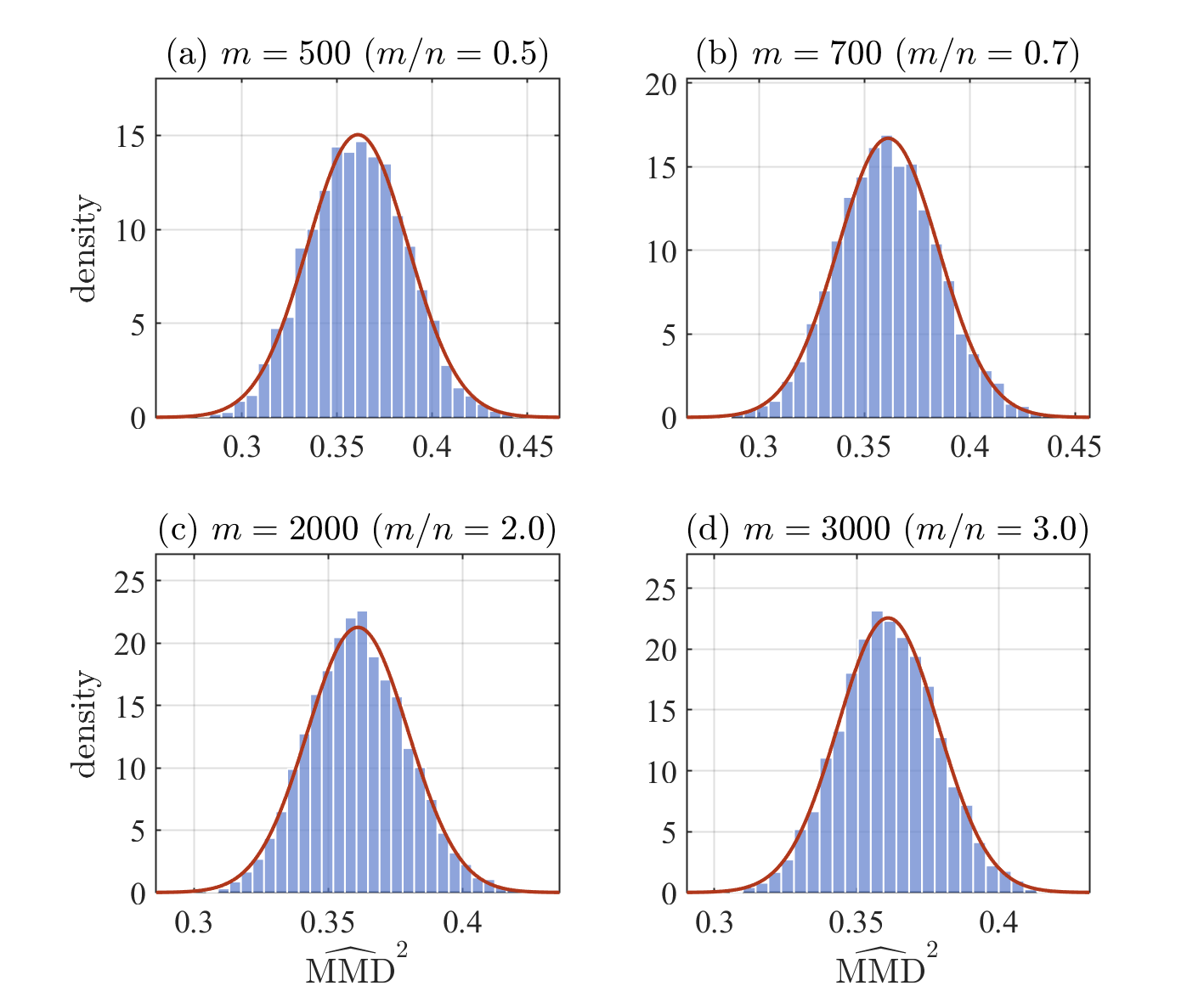}
	\caption{Empirical verification of asymptotic normality under unbalanced sampling.}
	\label{fig:normality}
\end{figure}

Figure~\ref{fig:normality} displays the empirical distributions of the estimated $\widehat{\mathrm{MMD}}^2$ across different sample size ratios, overlaid with the Gaussian density curves parameterized by the theoretical mean $\mathbb{E}[\widehat{\mathrm{MMD}}^2]$ and variance $\mathbb{E}[\hat V]$ derived from Corollary~\ref{thm:alt-dist}. As visually evident, the empirical histograms closely align with the theoretical density curves in all subfigures. This strong agreement validates our formulation, demonstrating that the sample limit ratios $\rho_X$ and $\rho_Y$ accurately capture the effect of sample imbalance on the asymptotic variance.

\subsection{Accuracy and efficiency of euMMD}

\paragraph{Statistical accuracy of euMMD}
To validate the statistical precision of the proposed accelerated euMMD estimator, we conduct a comparative analysis under the alternative hypothesis $\mathcal{H}_1$. We consider one-dimensional Laplace distributions where $\mathbb{P} \sim \mathrm{Laplace}(0,1)$ and $\mathbb{Q} \sim \mathrm{Laplace}(1,1)$, representing a location shift. To assess the performance under imbalanced conditions, the sample sizes are set at a ratio of $m = 1.2n$, with $n \in \{10, 100, 1000, 10000\}$. As a comparison baseline, we consider the FastMMD method based on Random Fourier Features (RFF) \cite{zhao2015fastmmd}, whose detailed theoretical derivations for the variance estimator are provided in Appendix~\ref{app3}. Being an approximation-based technique, the accuracy of FastMMD strongly depends on the number of random basis functions $L$.

As shown in Table~\ref{tab:sim_n10}--\ref{tab:sim_n10000}, the proposed euMMD statistic and its variance components ($\hat{V}_1$ and $\hat{V}_2$) are algebraically strictly identical to the classical matrix-form unbiased MMD estimator across all evaluated sample sizes. This perfect numerical alignment confirms that our $\mathcal{O}(N \log N)$ euMMD framework is not an approximation, but an exact algorithmic acceleration of the finite-sample U-statistic. In contrast, While FastMMD yields acceptable estimates for sufficiently large $L$ (e.g., $L=1024$), it intrinsically introduces approximation variance.

\begin{table}[!htb]
	\centering
	\caption{Comparison of MMD and variance estimators for $n=10, m=12$}
	\label{tab:sim_n10}
	\begin{tabular}{lcccc}
		\toprule
		Method & $\widehat{\mathrm{MMD}}^2$ & $\hat{V}$ (Total Var) & $\hat{V}_1$ (1st-order) & $\hat{V}_2$ (2nd-order) \\
		\midrule
		Exact (Matrix)     &  1.945e-03 & 6.234e-03 & 2.507e-03 & 3.727e-03 \\
		euMMD              &  1.945e-03 & 6.234e-03 & 2.507e-03 & 3.727e-03 \\
		FastMMD ($L=64$)   &  1.665e-02 & 4.571e-03 & 7.840e-04 & 3.787e-03 \\
		FastMMD ($L=128$)  &  3.957e-04 & 5.610e-03 & 1.210e-03 & 4.399e-03 \\
		FastMMD ($L=256$)  &  7.224e-04 & 6.459e-03 & 2.510e-03 & 3.950e-03 \\
		FastMMD ($L=512$)  &  1.910e-03 & 5.146e-03 & 1.767e-03 & 3.379e-03 \\
		FastMMD ($L=1024$) &  2.140e-03 & 5.569e-03 & 1.990e-03 & 3.579e-03 \\
		\bottomrule
	\end{tabular}
\end{table}

\begin{table}[!htb]
	\centering
	\caption{Comparison of MMD and variance estimators for $n=100, m=120$}
	\label{tab:sim_n100}
	\begin{tabular}{lcccc}
		\toprule
		Method & $\widehat{\mathrm{MMD}}^2$ & $\hat{V}$ (Total Var) & $\hat{V}_1$ (1st-order) & $\hat{V}_2$ (2nd-order) \\
		\midrule
		Exact (Matrix)     & 9.271e-02 & 1.142e-03 & 1.109e-03 & 3.206e-05 \\
		euMMD              & 9.271e-02 & 1.142e-03 & 1.109e-03 & 3.206e-05 \\
		FastMMD ($L=64$)   & 8.723e-02 & 1.058e-03 & 1.025e-03 & 3.339e-05 \\
		FastMMD ($L=128$)  & 9.376e-02 & 1.220e-03 & 1.182e-03 & 3.803e-05 \\
		FastMMD ($L=256$)  & 9.447e-02 & 1.198e-03 & 1.164e-03 & 3.455e-05 \\
		FastMMD ($L=512$)  & 8.719e-02 & 1.018e-03 & 9.882e-04 & 3.004e-05 \\
		FastMMD ($L=1024$) & 8.870e-02 & 1.055e-03 & 1.023e-03 & 3.115e-05 \\
		\bottomrule
	\end{tabular}
\end{table}

\begin{table}[!htb]
	\centering
	\caption{Comparison of MMD and variance estimators for $n=1,000, m=1,200$}
	\label{tab:sim_n1000}
	\begin{tabular}{lcccc}
		\toprule
		Method & $\widehat{\mathrm{MMD}}^2$ & $\hat{V}$ (Total Var) & $\hat{V}_1$ (1st-order) & $\hat{V}_2$ (2nd-order) \\
		\midrule
		Exact (Matrix)     & 1.042e-01 & 1.026e-04 & 1.023e-04 & 2.962e-07 \\
		euMMD              & 1.042e-01 & 1.026e-04 & 1.023e-04 & 2.962e-07 \\
		FastMMD ($L=64$)   & 9.925e-02 & 9.533e-05 & 9.502e-05 & 3.101e-07 \\
		FastMMD ($L=128$)  & 1.069e-01 & 1.111e-04 & 1.108e-04 & 3.550e-07 \\
		FastMMD ($L=256$)  & 1.075e-01 & 1.088e-04 & 1.085e-04 & 3.193e-07 \\
		FastMMD ($L=512$)  & 1.001e-01 & 9.336e-05 & 9.308e-05 & 2.785e-07 \\
		FastMMD ($L=1024$) & 1.011e-01 & 9.611e-05 & 9.582e-05 & 2.901e-07 \\
		\bottomrule
	\end{tabular}
\end{table}

\begin{table}[!htb]
	\centering
	\caption{Comparison of MMD and variance estimators for $n=10,000, m=12,000$}
	\label{tab:sim_n10000}
	\begin{tabular}{lcccc}
		\toprule
		Method & $\widehat{\mathrm{MMD}}^2$ & $\hat{V}$ (Total Var) & $\hat{V}_1$ (1st-order) & $\hat{V}_2$ (2nd-order) \\
		\midrule
		Exact (Matrix)     & 1.144e-01 & 1.147e-05 & 1.147e-05 & 2.939e-09 \\
		euMMD              & 1.144e-01 & 1.147e-05 & 1.147e-05 & 2.939e-09 \\
		FastMMD ($L=64$)   & 1.103e-01 & 1.081e-05 & 1.081e-05 & 3.090e-09 \\
		FastMMD ($L=128$)  & 1.178e-01 & 1.254e-05 & 1.253e-05 & 3.534e-09 \\
		FastMMD ($L=256$)  & 1.178e-01 & 1.214e-05 & 1.213e-05 & 3.166e-09 \\
		FastMMD ($L=512$)  & 1.098e-01 & 1.039e-05 & 1.039e-05 & 2.758e-09 \\
		FastMMD ($L=1024$) & 1.112e-01 & 1.076e-05 & 1.076e-05 & 2.881e-09 \\
		\bottomrule
	\end{tabular}
\end{table}

\begin{figure}[htb]
	\centering
	\includegraphics{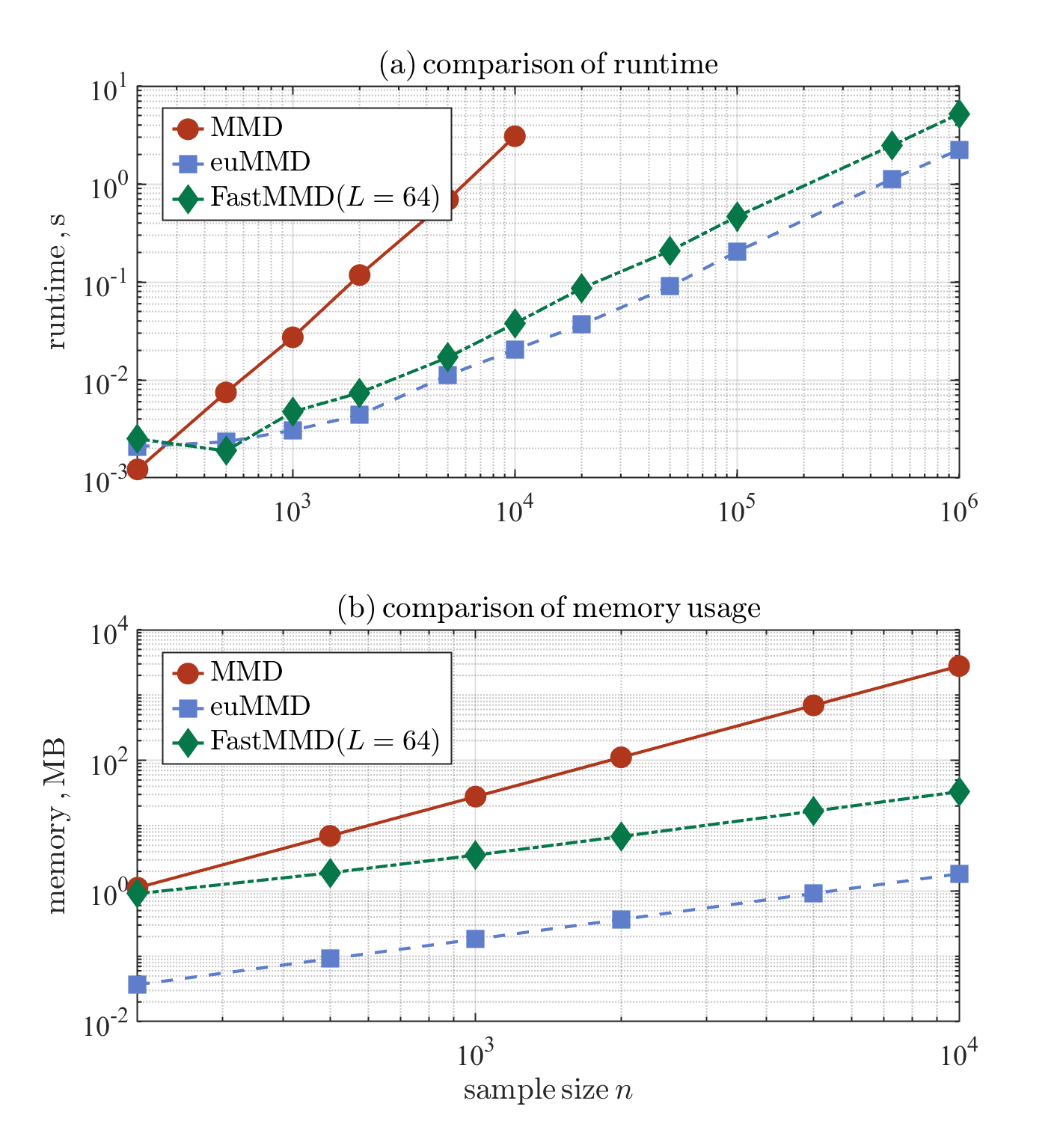}
	\caption{Double-logarithmic comparison of runtime and memory consumption for algorithms under unequal sample sizes.}
	\label{fig:unbalance_eff}
\end{figure}

\paragraph{Efficiency of variance estimation}
To evaluate the computational efficiency and scalability of the proposed accelerated framework for large-scale variance estimation, we compare its runtime and memory consumption against both the conventional $\mathcal{O}(N^2)$ algorithm and the FastMMD algorithm. Operating in a $2L$-dimensional Random Fourier Feature space, FastMMD computes the unbiased MMD and its variance estimator with a time complexity of $\mathcal{O}(NLd + NL^2)$ and a space complexity of $\mathcal{O}(NL)$, where $d$ denotes the original feature dimension. In the univariate case ($d=1$), the variance estimation step dominates the computational cost, resulting in an overall time complexity of $\mathcal{O}(NL^3)$. Due to the quadratic memory requirement, the conventional method is restricted to sample sizes $n \le 10^4$, whereas the proposed method scales up to $n = 10^6$. All experiments are independently repeated 20 times, and the averaged results are shown in Figure~\ref{fig:unbalance_eff}.

Under the double-logarithmic scale, clear separations between the three methods are observed in both runtime and memory usage. The conventional kernel-matrix-based algorithm exhibits quadratic growth ($\mathcal{O}(N^2)$), while the proposed method shows a significantly milder trend consistent with its $\mathcal{O}(N \log N)$ complexity. Compared with FastMMD, the proposed method achieves consistently lower runtime across all scales. Although the gap is moderate under the current setting with $L=64$, it is expected to widen in practice as higher approximation accuracy typically requires larger $L$, increasing the computational cost of FastMMD. Notably, even at the million-sample scale ($n = 10^6$), the proposed method remains highly efficient, completing variance estimation within seconds, whereas the conventional approach becomes computationally infeasible.

A similar but more pronounced gap is observed in memory consumption. The conventional method relies on explicit Gram matrix construction, leading to $\mathcal{O}(N^2)$ space complexity and reaching approximately 2.8~GB at $n=10{,}000$, becoming infeasible due to out-of-memory constraints. FastMMD reduces this requirement to 33.15~MB by operating in a randomized feature space. In contrast, the proposed method avoids explicit matrix or feature-map materialization and requires only $\mathcal{O}(N)$ auxiliary memory, with a peak usage of 1.83~MB at $n=10{,}000$, achieving over 1500 times reduction compared to the conventional method and about 18 times reduction compared to FastMMD. This compact memory footprint enables exact variance estimation at scales beyond the reach of standard implementations.

It is worth noting that FastMMD is particularly effective in high-dimensional settings due to its random Fourier feature representation; however, such extensions are beyond the scope of this work and are left for future investigation.

\subsection{Monitoring TimeGAN for ECG synthesis}

To demonstrate our extended euMMD framework, we monitor the training dynamics of a Time-series Generative Adversarial Network (TimeGAN) synthesizing electrocardiogram (ECG) signals\cite{yoon2019time} .\footnote{Code adapted from \url{https://github.com/HelenaFP/TimeGAN-for-ecgs}.} The model trains on real-world data from the MIT-BIH Arrhythmia Database \cite{moody2001impact}.\footnote{Available at \url{https://www.physionet.org/content/mitdb/1.0.0/}.}

\begin{figure}[htb]
	\centering
	\includegraphics{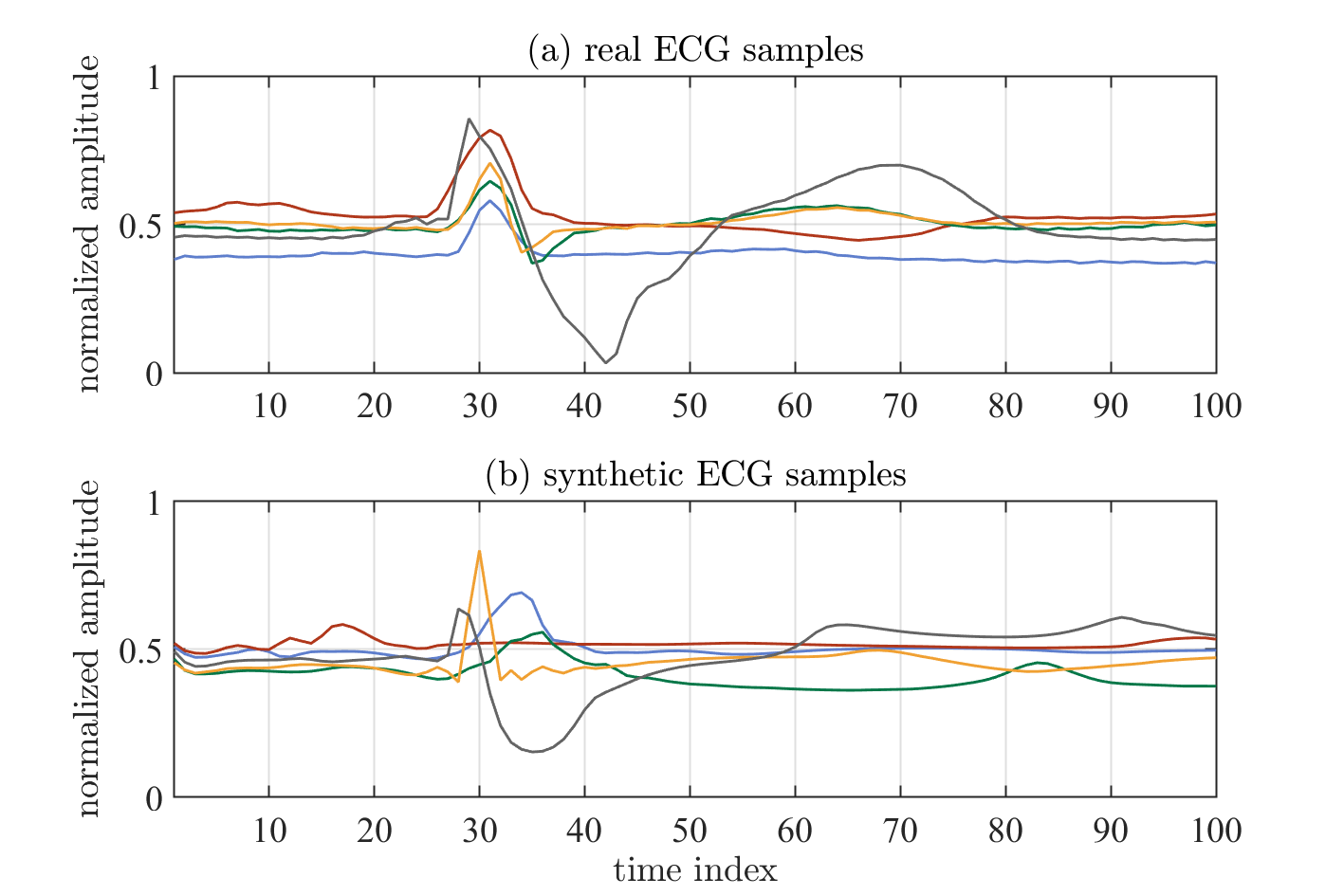}
	\caption{Comparison of normalized amplitude trajectories between (a) real ECG samples from the MIT-BIH database and (b) synthetic ECG samples generated by TimeGAN at iteration 20,000.}
	\label{fig:ecg_samples}
\end{figure}

During preprocessing, heartbeats are segmented and phase-normalized, yielding $n=109,398$ real samples. As illustrated in Figure~\ref{fig:ecg_samples}, after 20,000 iterations, the trained TimeGAN generates synthetic ECG signals with high morphological consistency and strong visual fidelity to the real patient data.

\begin{figure}[htb]
	\centering
	\includegraphics{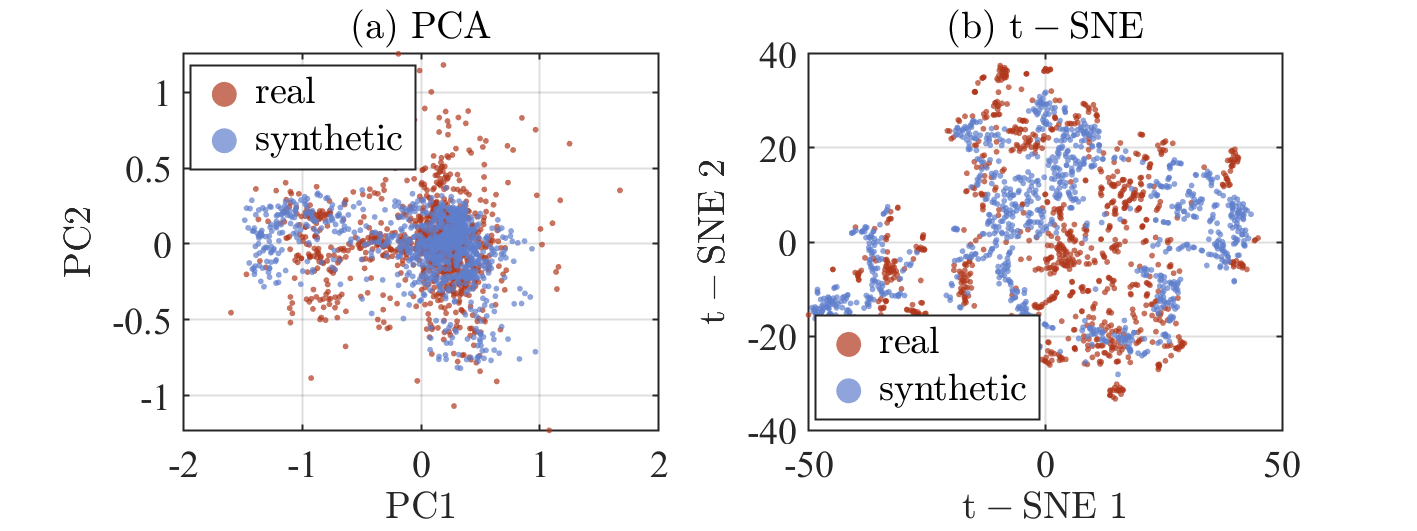}
	\caption{Two-dimensional latent space visualizations of real and synthetic ECG distributions at iteration 16,000 using (a) PCA and (b) t-SNE.}
	\label{fig:pca_tsne}
\end{figure}

Traditionally, generative fidelity is further evaluated heuristically using visual embeddings such as PCA and t-SNE \cite{hotelling1933analysis, van2008visualizing}. To contextualize our experiments, we project the high-dimensional sequences at iteration 16,000 into a two-dimensional space (Figure~\ref{fig:pca_tsne}). While their overlapping scatter plots qualitatively confirm global distributional alignment, such visual inspections remain highly subjective.

To overcome the limitations of purely qualitative assessments, a rigorous quantitative metric is required. However, monitoring distributional convergence over tens of thousands of training iterations is typically computationally prohibitive for standard MMD estimators due to massive memory footprints. Leveraging our proposed euMMD framework solves this bottleneck by efficiently computing the exact statistic and its unbiased variance. To manage the high dimensionality of the ECG signals, we evaluate convergence on their first two principal components, comparing the $n$ real samples against $m$ on-demand synthetic samples.

\begin{figure}[htb]
	\centering
	\includegraphics{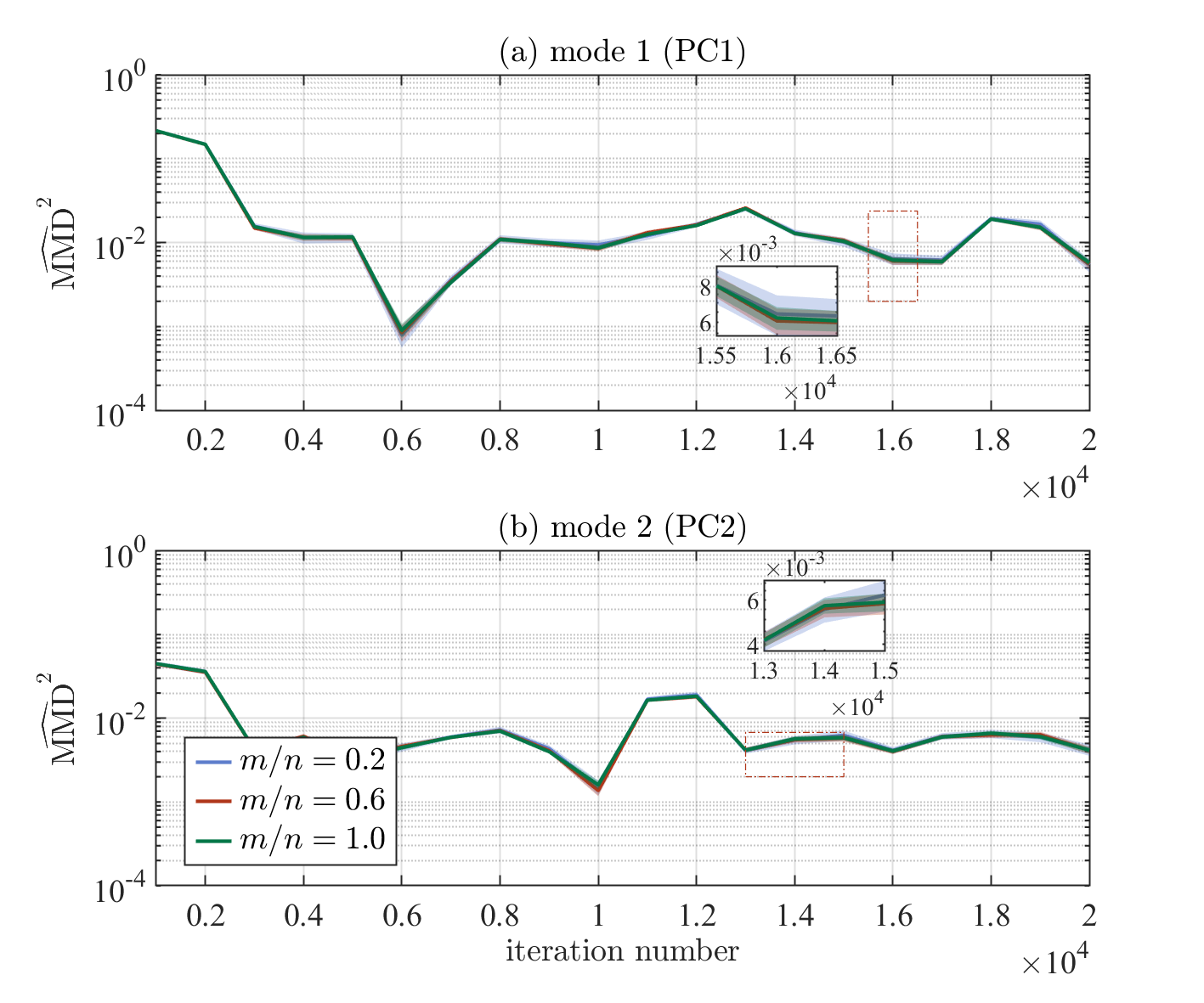}
	\caption{The trajectory of $\widehat{\mathrm{MMD}}^2$ and its $3\sigma$ confidence bounds over training iterations for the first two principal components (PC1 and PC2). Different colors represent varying sample ratios ($m/n$).}
	\label{fig:mmd_convergence}
\end{figure}

Figure~\ref{fig:mmd_convergence} presents the convergence behavior of euMMD along with its $3\sigma$ confidence intervals under different sample ratios ($m/n \in \{0.2, 0.6, 1.0\}$). The corresponding average computation times are reported in Table~\ref{tab:comp_time}. As shown, the classical matrix-based implementation becomes computationally infeasible at large scale ($n \approx 10^5$), leading to out-of-memory issues, whereas the proposed accelerated algorithm (euMMD) is able to complete the computation within tens of milliseconds (approximately 22.8 ms to 38.7 ms).

Moreover, the empirical results suggest that the MMD estimator remains unbiased and consistent across all tested ratios, while a smaller synthetic sample size ($m$) leads to a moderate increase in estimation variance. These observations indicate a practical trade-off between computational efficiency and estimation accuracy: reducing the synthetic sample ratio can further improve computational speed under the quasi-linear algorithmic framework, at the cost of a slightly increased estimation variance.

\begin{table}[htb]
	\centering
	\caption{Detailed computation time of the euMMD algorithm across all TimeGAN iterations (in seconds).}
	\label{tab:comp_time}
		\begin{tabular}{lcccccc}
			\toprule
			\multirow{3}{*}{Iteration} & \multicolumn{2}{c}{Ratio $= 0.2$} & \multicolumn{2}{c}{Ratio $= 0.6$} & \multicolumn{2}{c}{Ratio $= 1.0$} \\
			\cmidrule(lr){2-3} \cmidrule(lr){4-5} \cmidrule(lr){6-7}
			& PC1& PC2& PC1& PC2 & PC1 & PC2 \\
			\midrule
			1000  & 1.992e-02 & 2.059e-02 & 2.714e-02 & 3.015e-02 & 3.309e-02 & 3.636e-02 \\
			2000  & 1.990e-02 & 2.062e-02 & 2.766e-02 & 2.993e-02 & 3.737e-02 & 3.689e-02 \\
			3000  & 2.079e-02 & 2.075e-02 & 2.980e-02 & 2.948e-02 & 3.715e-02 & 3.726e-02 \\
			4000  & 2.089e-02 & 2.050e-02 & 2.927e-02 & 2.981e-02 & 3.901e-02 & 3.792e-02 \\
			5000  & 2.036e-02 & 2.129e-02 & 2.969e-02 & 2.943e-02 & 3.833e-02 & 3.832e-02 \\
			6000  & 2.421e-02 & 2.851e-02 & 3.055e-02 & 2.970e-02 & 4.217e-02 & 3.954e-02 \\
			7000  & 2.109e-02 & 2.007e-02 & 2.969e-02 & 3.077e-02 & 3.834e-02 & 3.740e-02 \\
			8000  & 2.641e-02 & 2.065e-02 & 2.985e-02 & 2.937e-02 & 3.904e-02 & 3.823e-02 \\
			9000  & 2.149e-02 & 2.086e-02 & 2.925e-02 & 2.943e-02 & 3.888e-02 & 3.788e-02 \\
			10000 & 7.222e-02 & 4.194e-02 & 3.635e-02 & 3.182e-02 & 6.294e-02 & 3.874e-02 \\
			11000 & 2.088e-02 & 2.042e-02 & 2.927e-02 & 2.858e-02 & 3.841e-02 & 3.680e-02 \\
			12000 & 2.054e-02 & 2.054e-02 & 2.882e-02 & 2.877e-02 & 3.774e-02 & 3.705e-02 \\
			13000 & 2.056e-02 & 2.019e-02 & 2.922e-02 & 3.011e-02 & 3.761e-02 & 3.877e-02 \\
			14000 & 2.064e-02 & 2.025e-02 & 2.924e-02 & 2.956e-02 & 3.724e-02 & 3.735e-02 \\
			15000 & 2.106e-02 & 2.007e-02 & 2.952e-02 & 3.014e-02 & 3.781e-02 & 3.895e-02 \\
			16000 & 2.031e-02 & 2.044e-02 & 2.999e-02 & 3.065e-02 & 4.092e-02 & 3.936e-02 \\
			17000 & 2.074e-02 & 2.070e-02 & 3.153e-02 & 2.938e-02 & 3.879e-02 & 3.763e-02 \\
			18000 & 2.071e-02 & 2.004e-02 & 2.997e-02 & 2.945e-02 & 3.755e-02 & 3.809e-02 \\
			19000 & 2.092e-02 & 2.026e-02 & 2.958e-02 & 2.924e-02 & 3.870e-02 & 3.754e-02 \\
			20000 & 2.053e-02 & 2.017e-02 & 2.953e-02 & 2.988e-02 & 3.905e-02 & 3.918e-02 \\
			\midrule
			Average & \multicolumn{2}{c}{2.283e-02} & \multicolumn{2}{c}{2.979e-02} & \multicolumn{2}{c}{3.874e-02} \\
			\bottomrule
		\end{tabular}
\end{table}

Crucially, equipped with these exact variance estimates, one can readily formulate robust stopping criteria. For instance, a \textit{confidence-bounded threshold} ensures the upper confidence limit falls below a predefined tolerance $\epsilon$:
\begin{equation}
	\widehat{\mathrm{MMD}}^2 + 3\sqrt{\widehat{V}} < \epsilon.
\end{equation}
Other statistical decision rules can be analogously derived, though they remain outside our primary scope.

\section{Conclusion}
\label{sec5}

In this paper, we address the theoretical and computational bottlenecks of Maximum Mean Discrepancy (MMD) variance estimation, yielding the following main contributions:

\begin{itemize}
	\item[(1)] A unbiased finite-sample MMD variance estimator is derived for unbalanced regimes. By correcting inconsistencies in previous combinatorial formulations, the resulting estimator provides reliable estimation under both null and alternative hypotheses without introducing approximation bias.
	
	\item[(2)] A generalized prefix-suffix accumulator framework is developed for the Laplace kernel. This algorithmic construction reduces the time complexity from $\mathcal{O}(N^2)$ to $\mathcal{O}(N \log N)$ and the space complexity to $\mathcal{O}(N)$, while preserving exactness and avoiding kernel-matrix construction.
	
	\item[(3)] Empirical validation confirms the theoretical properties of the proposed estimators, verifying their strict algebraic equivalence to classical matrix-based formulations without introducing approximation bias.
	
	\item[(4)] Computational benchmarks demonstrate superior memory efficiency and scalability compared to approximation methods like FastMMD, enabling practical large-scale applications such as efficiently monitoring the training dynamics of TimeGAN for ECG synthesis.
	
	\item[(5)] Future work includes extending the recursive accumulator framework to broader classes of shift-invariant kernels and exploring adaptive bandwidth selection for high-dimensional generative modeling.
\end{itemize}

\backmatter

\bmhead{Supplementary information}

A supplementary zip file is provided including the core MATLAB implementations of the proposed unified MMD variance estimator, the $O(N \log N)$ acceleration engine (euMMD), and the scripts for reproducing numerical experiments.

\bmhead{Acknowledgements}

The authors thank Prof. Pengfei Wei (Northwestern Polytechnical University) for the insightful discussions and for proofreading the manuscript.

\section*{Declarations}

\begin{itemize}
	\item Funding: This work was supported by the Defense Industrial Technology Development Program (Grant No. JCKY2022607C002).
	\item Conflict of interest/Competing interests:  The authors have no relevant financial or non-financial interests to disclose.
	\item Ethics approval and consent to participate: Not applicable.
	\item Consent for publication: Not applicable.
	\item Data availability: The datasets are available at \url{https://drive.google.com/drive/folders/1dvNL_4IrQy2qhGbIHOOei-XDRoQUgoFs?usp=drive_link}.
	\item Materials availability: Not applicable.
	\item Code availability: The code is available at \url{https://github.com/shijiezhong19-star/Unified-MMD-Stat}. 
	\item Author contribution: S.Z. and J.F. conceived the study and methodology. S.Z. established the theoretical framework, performed mathematical derivations, and developed the accelerated algorithms. Y.Y. and D.G. assisted with numerical execution. J.F. provided overall supervision and guidance. S.Z. and Y.Y. prepared the original draft. All authors reviewed and approved the final manuscript.
\end{itemize}

\begin{appendices}
\section{Revisiting the derivation of MMD variance}
\label{App1}
To rigorously derive the exact variance of the MMD estimator in the unbalanced regime ($n \neq m$), we rely on the theory of generalized U-statistics \cite{sen1974weak}. Since MMD is a generalized two-sample U-statistic of degrees $(2, 2)$, its exact variance can be directly expanded as a combinatorial sum of conditional variances $\zeta_{c, d}$ spanning interaction orders 1 through 4 (i.e., for $c, d \in \{0, 1, 2\}$):
\begin{equation}
	\mathrm{Var}(\widehat{\mathrm{MMD}}^2) = \sum_{c=0}^2 \sum_{d=0}^2 \binom{n}{2}^{-1} \binom{2}{c} \binom{n-2}{2-c} \binom{m}{2}^{-1} \binom{2}{d} \binom{m-2}{2-d} \zeta_{c, d},
	\label{eq:mmd-var-combinatorial}
\end{equation}
where $\zeta_{0,0} = 0$, and the specific conditional variances are defined as:
\begin{equation}
	\zeta_{c, d} = \mathrm{Var}\big(\mathbb{E}[h(\mathbf{X}, \mathbf{Y}) \mid X_1,\dots, X_c, Y_1,\dots, Y_d]\big).
\end{equation}

To streamline the notation for these components, we let $\Delta\mu = \mu_{\mathbb{P}} - \mu_{\mathbb{Q}}$ and define $\nu_{\mathbb{P}} := \langle \Delta\mu, C_{\mathbb{P}} \Delta\mu \rangle_{\mathcal{H}}$ and $\nu_{\mathbb{Q}} := \langle \Delta\mu, C_{\mathbb{Q}} \Delta\mu \rangle_{\mathcal{H}}$. As recently shown by Wei et al. \cite{wei2025maximum}, most of these conditional variances can be evaluated symmetrically without complication. Specifically, the terms spanning interaction orders 1 through 4 (excluding the cross-sample term $\zeta_{XY}$) are correctly given by:
\begin{gather}
	\zeta_X = \nu_{\mathbb{P}},
	\qquad
	\zeta_Y = \nu_{\mathbb{Q}}, \nonumber\\
	\zeta_{XX'} = \|C_{\mathbb{P}}\|^2_{\mathrm{HS}} + 2 \nu_{\mathbb{P}},
	\qquad
	\zeta_{YY'} = \|C_{\mathbb{Q}}\|^2_{\mathrm{HS}} + 2 \nu_{\mathbb{Q}},
	\nonumber\\
	\zeta_{XX'Y} = \|C_{\mathbb{P}}\|_{\mathrm{HS}}^2 + \frac{1}{2} \langle C_{\mathbb{P}}, C_{\mathbb{Q}} \rangle_{\mathrm{HS}} + 2 \nu_{\mathbb{P}} + \nu_{\mathbb{Q}}, \nonumber\\
	\zeta_{XYY'} = \|C_{\mathbb{Q}}\|_{\mathrm{HS}}^2 + \frac{1}{2} \langle C_{\mathbb{P}}, C_{\mathbb{Q}} \rangle_{\mathrm{HS}} + \nu_{\mathbb{P}} + 2 \nu_{\mathbb{Q}}, \nonumber\\
	\zeta_{XX'YY'} = \|C_{\mathbb{P}}\|_{\mathrm{HS}}^2 + \|C_{\mathbb{Q}}\|_{\mathrm{HS}}^2 + \langle C_{\mathbb{P}}, C_{\mathbb{Q}} \rangle_{\mathrm{HS}} + 2 \nu_{\mathbb{P}} + 2 \nu_{\mathbb{Q}}.
\end{gather}

However, the direct finite-sample expansion proposed in \cite{wei2025maximum} contains a subtle algebraic inconsistency specifically in the remaining cross-sample projection term $\zeta_{XY}$, resulting in an entangled overall variance formulation.

\subsection{Correction of the cross-sample interaction}

To correct this, we focus on the proper derivation of $\zeta_{XY}$. The conditional expectation with respect to one sample from each distribution is:
\begin{equation*}
	\mathbb{E}[h(X,X';Y,Y') \mid X,Y] = \mu_{\mathbb{P}}(X) + \mu_{\mathbb{Q}}(Y) - \frac{1}{2} \left[\mu_{\mathbb{Q}}(X) + \mu_{\mathbb{P}}(Y) + k(X,Y) + \mathbb{E}[k(X,Y)] \right].
\end{equation*}
Noting that $\langle C_{\mathbb{P}}, C_{\mathbb{Q}} \rangle_{\mathrm{HS}} = \mathrm{Var}(k(X,Y)) - \langle \mu_{\mathbb{Q}}, C_{\mathbb{P}} \mu_{\mathbb{Q}} \rangle - \langle \mu_{\mathbb{P}}, C_{\mathbb{Q}} \mu_{\mathbb{P}} \rangle$, we can expand $\zeta_{XY}$ as:
\begin{align}
	\zeta_{XY} &= \mathrm{Var} \left(\mu_{\mathbb{P}}(X) - \frac{1}{2}\mu_{\mathbb{Q}}(X) \right) + \mathrm{Var}\left(\mu_{\mathbb{Q}}(Y) - \frac{1}{2}\mu_{\mathbb{P}}(Y) \right) + \frac{1}{4}\mathrm{Var}(k(X,Y)) \nonumber\\
	&\quad \boldsymbol{-} \mathrm{Cov}\left(\mu_{\mathbb{P}}(X) - \frac{1}{2}\mu_{\mathbb{Q}}(X), k(X,Y) \right) \boldsymbol{-} \mathrm{Cov}\left(\mu_{\mathbb{Q}}(Y) - \frac{1}{2}\mu_{\mathbb{P}}(Y), k(X,Y) \right) \nonumber\\
	&= \langle \mu_{\mathbb{P}} - \frac{1}{2}\mu_{\mathbb{Q}}, C_{\mathbb{P}}\left(\mu_{\mathbb{P}} - \frac{1}{2}\mu_{\mathbb{Q}}\right)\rangle + \langle \mu_{\mathbb{Q}} - \frac{1}{2}\mu_{\mathbb{P}}, C_{\mathbb{Q}}\left(\mu_{\mathbb{Q}} - \frac{1}{2}\mu_{\mathbb{P}}\right)\rangle \nonumber\\
	&\quad + \frac{1}{4} \langle C_{\mathbb{P}}, C_{\mathbb{Q}} \rangle_{\mathrm{HS}} + \frac{1}{4} \langle \mu_{\mathbb{Q}}, C_{\mathbb{P}} \mu_{\mathbb{Q}} \rangle + \frac{1}{4} \langle \mu_{\mathbb{P}}, C_{\mathbb{Q}} \mu_{\mathbb{P}} \rangle \nonumber\\
	&\quad \boldsymbol{-} \langle \mu_{\mathbb{P}} - \frac{1}{2} \mu_{\mathbb{Q}}, C_{\mathbb{P}} \mu_{\mathbb{Q}} \rangle \boldsymbol{-} \langle \mu_{\mathbb{Q}} - \frac{1}{2} \mu_{\mathbb{P}}, C_{\mathbb{Q}} \mu_{\mathbb{P}} \rangle \nonumber\\
	&= \frac{1}{4} \langle C_{\mathbb{P}}, C_{\mathbb{Q}} \rangle_{\mathrm{HS}} + \langle \mu_{\mathbb{P}} - \mu_{\mathbb{Q}}, C_{\mathbb{P}}(\mu_{\mathbb{P}} - \mu_{\mathbb{Q}}) \rangle + \langle \mu_{\mathbb{P}} - \mu_{\mathbb{Q}}, C_{\mathbb{Q}}(\mu_{\mathbb{P}} - \mu_{\mathbb{Q}}) \rangle \nonumber\\
	&= \frac{1}{4} \langle C_{\mathbb{P}}, C_{\mathbb{Q}} \rangle_{\mathrm{HS}} + \nu_{\mathbb{P}} + \nu_{\mathbb{Q}},
\end{align}
where the last line follows by grouping and algebraically simplifying the components for each covariance operator. For instance, isolating the $C_{\mathbb{P}}$ terms yields:
\begin{align}
	&\quad\langle \mu_{\mathbb{P}} - \frac{1}{2}\mu_{\mathbb{Q}}, C_{\mathbb{P}}(\mu_{\mathbb{P}} - \frac{1}{2}\mu_{\mathbb{Q}}) \rangle + \frac{1}{4} \langle \mu_{\mathbb{Q}}, C_{\mathbb{P}} \mu_{\mathbb{Q}} \rangle - \langle \mu_{\mathbb{P}} - \frac{1}{2} \mu_{\mathbb{Q}}, C_{\mathbb{P}} \mu_{\mathbb{Q}} \rangle \nonumber \\
	&= \left( \langle \mu_{\mathbb{P}}, C_{\mathbb{P}} \mu_{\mathbb{P}} \rangle - \langle \mu_{\mathbb{P}}, C_{\mathbb{P}} \mu_{\mathbb{Q}} \rangle + \frac{1}{4}\langle \mu_{\mathbb{Q}}, C_{\mathbb{P}} \mu_{\mathbb{Q}} \rangle \right) + \frac{1}{4} \langle \mu_{\mathbb{Q}}, C_{\mathbb{P}} \mu_{\mathbb{Q}} \rangle - \left( \langle \mu_{\mathbb{P}}, C_{\mathbb{P}} \mu_{\mathbb{Q}} \rangle - \frac{1}{2}\langle \mu_{\mathbb{Q}}, C_{\mathbb{P}} \mu_{\mathbb{Q}} \rangle \right) \nonumber\\
	&= \langle \mu_{\mathbb{P}}, C_{\mathbb{P}} \mu_{\mathbb{P}} \rangle - \langle \mu_{\mathbb{P}}, C_{\mathbb{P}} \mu_{\mathbb{Q}} \rangle + \frac{1}{4}\langle \mu_{\mathbb{Q}}, C_{\mathbb{P}} \mu_{\mathbb{Q}} \rangle + \frac{1}{4} \langle \mu_{\mathbb{Q}}, C_{\mathbb{P}} \mu_{\mathbb{Q}} \rangle - \langle \mu_{\mathbb{P}}, C_{\mathbb{P}} \mu_{\mathbb{Q}} \rangle + \frac{1}{2}\langle \mu_{\mathbb{Q}}, C_{\mathbb{P}} \mu_{\mathbb{Q}} \rangle \nonumber\\
	&= \langle \mu_{\mathbb{P}}, C_{\mathbb{P}} \mu_{\mathbb{P}} \rangle - 2\langle \mu_{\mathbb{P}}, C_{\mathbb{P}} \mu_{\mathbb{Q}} \rangle + \langle \mu_{\mathbb{Q}}, C_{\mathbb{P}} \mu_{\mathbb{Q}} \rangle \nonumber\\
	&= \langle \mu_{\mathbb{P}} - \mu_{\mathbb{Q}}, C_{\mathbb{P}}(\mu_{\mathbb{P}} - \mu_{\mathbb{Q}}) \rangle.
	\label{V1operate}
\end{align}
A completely symmetric reduction applies to the $C_{\mathbb{Q}}$ components. Crucially, as highlighted in the derivation above, the covariance cross-terms must carry a negative sign. This proper tracking of the negative signs is exactly where the direct expansion in \cite{wei2025maximum} diverges.

\subsection{Algebraic synthesis and simplification}
To understand how the formulation condenses into the decoupled exact variance, we explicitly write the fully expanded weighted sum. Substituting the corrected $\zeta$ values back into Eq.~\eqref{eq:mmd-var-combinatorial}, the full expression is:
\begin{multline}
	\mathrm{Var}(\widehat{\mathrm{MMD}}^2) =
	\frac{4}{n} \left( \frac{n-2}{n-1} \cdot \frac{m-2}{m} \cdot \frac{m-3}{m-1} \right) \nu_{\mathbb{P}}
	+ \frac{4}{m} \left( \frac{n-2}{n} \cdot \frac{n-3}{n-1} \cdot \frac{m-2}{m-1} \right) \nu_{\mathbb{Q}}
	\nonumber\\
	+ \frac{2}{n (n - 1)} \left( \frac{m-2}{m} \cdot \frac{m-3}{m-1} \right) (\|C_{\mathbb{P}}\|_{\mathrm{HS}}^2 + 2 \nu_{\mathbb{P}})
	+ \frac{2}{m (m-1)} \left( \frac{n-2}{n} \cdot \frac{n-3}{n-1} \right) (\|C_{\mathbb{Q}}\|_{\mathrm{HS}}^2 + 2 \nu_{\mathbb{Q}})
	\nonumber\\
	+ \frac{16}{n m} \left( \frac{n-2}{n-1} \cdot \frac{m-2}{m-1} \right) \left[ \frac{1}{4} \langle C_{\mathbb{P}}, C_{\mathbb{Q}} \rangle_{\mathrm{HS}} + \nu_{\mathbb{P}} + \nu_{\mathbb{Q}} \right]
	\nonumber\\
	+ \frac{8}{n (n-1) m} \left( \frac{m-2}{m-1} \right)
	\left[ \|C_{\mathbb{P}}\|_{\mathrm{HS}}^2 + \frac{1}{2} \langle C_{\mathbb{P}}, C_{\mathbb{Q}} \rangle_{\mathrm{HS}} + 2 \nu_{\mathbb{P}} + \nu_{\mathbb{Q}} \right]
	\nonumber\\
	+ \frac{8}{n m (m - 1)} \left( \frac{n-2}{n-1} \right) \left[ \|C_{\mathbb{Q}}\|_{\mathrm{HS}}^2 + \frac{1}{2} \langle C_{\mathbb{P}}, C_{\mathbb{Q}} \rangle_{\mathrm{HS}} + \nu_{\mathbb{P}} + 2 \nu_{\mathbb{Q}} \right]
	\nonumber\\
	+ \frac{4}{n(n-1) m(m-1)}
	\left[ \|C_{\mathbb{P}}\|_{\mathrm{HS}}^2 + \|C_{\mathbb{Q}}\|_{\mathrm{HS}}^2 + \langle C_{\mathbb{P}}, C_{\mathbb{Q}} \rangle_{\mathrm{HS}} + 2 \nu_{\mathbb{P}} + 2 \nu_{\mathbb{Q}} \right].
\end{multline}

Despite the initial combinatorial complexity, correcting the cross-terms ensures a systematic and exact cancellation across all components. By extracting the common denominator $n(n-1)m(m-1)$, we can aggregate the coefficients for each specific functional term.

For the first-order term $\nu_{\mathbb{P}}$, gathering its coefficients from across the expansion yields:
\begin{align*}
	&\quad\frac{4}{n(n-1)m(m-1)} \Big[ (n-2)(m-2)(m-3) + (m-2)(m-3) \\
	&\qquad + 4(n-2)(m-2) + 4(m-2) + 2(n-2) + 2 \Big] \\
	&= \frac{4(n-1)}{n(n-1)m(m-1)} \Big[ m^2 - 5m + 6 + 4m - 8 + 2 \Big] \\
	&= \frac{4(n-1)}{n(n-1)m(m-1)} \big[ m(m-1) \big] = \frac{4}{n}.
\end{align*}
By perfect symmetry with respect to sample sizes $n$ and $m$, gathering the coefficients for $\nu_{\mathbb{Q}}$ symmetrically simplifies to:
\begin{equation*}
	\frac{4(m-1)}{n(n-1)m(m-1)} \big[ n(n-1) \big] = \frac{4}{m}.
\end{equation*}

For the second-order term $\|C_{\mathbb{P}}\|_{\mathrm{HS}}^2$, aggregating its components directly collapses the interacting polynomial:
\begin{align*}
	&\quad\frac{2}{n(n-1)m(m-1)} \Big[ (m-2)(m-3) + 4(m-2) + 2 \Big] \\
	&= \frac{2}{n(n-1)m(m-1)} \Big[ m^2 - 5m + 6 + 4m - 8 + 2 \Big] \\
	&= \frac{2}{n(n-1)m(m-1)} \big[ m(m-1) \big] = \frac{2}{n(n-1)}.
\end{align*}
Correspondingly, gathering the $\|C_{\mathbb{Q}}\|_{\mathrm{HS}}^2$ components yields an identical symmetric cancellation, reducing exactly to:
\begin{equation*}
	\frac{2}{n(n-1)m(m-1)} \big[ n(n-1) \big] = \frac{2}{m(m-1)}.
\end{equation*}

Finally, aggregating the cross-covariance $\langle C_{\mathbb{P}}, C_{\mathbb{Q}} \rangle_{\mathrm{HS}}$ terms efficiently factors the remaining elements:
\begin{align*}
	&\quad\frac{4}{n(n-1)m(m-1)} \Big[ (n-2)(m-2) + (m-2) + (n-2) + 1 \Big] \\
	&= \frac{4}{n(n-1)m(m-1)} \Big[ (n-2+1)(m-2+1) \Big] \\
	&= \frac{4}{n(n-1)m(m-1)} \big[ (n-1)(m-1) \big] = \frac{4}{nm}.
\end{align*}

Consequently, aggregating all five fully simplified components exactly recovers the clean, decoupled variance framework:
\begin{equation}
	\mathrm{Var}(\widehat{\mathrm{MMD}}^2) = \frac{4}{n}\nu_{\mathbb{P}} + \frac{4}{m}\nu_{\mathbb{Q}} + \frac{2}{n(n-1)}\|C_{\mathbb{P}}\|_{\mathrm{HS}}^2 + \frac{2}{m(m-1)}\|C_{\mathbb{Q}}\|_{\mathrm{HS}}^2 + \frac{4}{nm}\langle C_{\mathbb{P}}, C_{\mathbb{Q}} \rangle_{\mathrm{HS}}.
\end{equation}

\section{Review and extension of unbiased sub-estimators}
\label{App2}

The principle of unbiased estimation for U-statistics requires excluding all summation terms involving repeated data indices. We summarize below the matrix-form estimators of the key components. The product of mean embedding terms and squared kernel terms follow the definitions of Sutherland et al.~\cite{sutherland2019unbiased}. However, for the quadratic forms involving covariance operators, we re-derive the estimators using centered covariance operators, whereas Sutherland et al. employ the uncentered formulation.

\subsection{Products of mean embeddings}

\begin{enumerate}
	\item \textbf{squared cross-product}:
	\begin{align}
		\widehat{\langle \mu_{\mathbb{P}}, \mu_{\mathbb{Q}} \rangle^2_{\mathcal{H}}} = \frac{1}{(n)_2 (m)_2} \left[ (\mathbf{1}_n^\top \mathbf{K}_{\mathbf{XY}} \mathbf{1}_m)^2 - \|\mathbf{K}_{\mathbf{XY}}^\top \mathbf{1}_n\|_2^2 - \|\mathbf{K}_{\mathbf{XY}} \mathbf{1}_m\|_2^2 + \|\mathbf{K}_{\mathbf{XY}}\|_F^2 \right].
	\end{align}
	
	\item \textbf{squared within-products}:
	\begin{align}
		\widehat{\langle \mu_{\mathbb{P}}, \mu_{\mathbb{P}} \rangle^2_{\mathcal{H}}} &= \frac{1}{(n)_4} \left[ (\mathbf{1}_n^\top \tilde{\mathbf{K}}_{\mathbf{XX}} \mathbf{1}_n)^2 - 4 \|\tilde{\mathbf{K}}_{\mathbf{XX}} \mathbf{1}_n\|_2^2 + 2 \|\tilde{\mathbf{K}}_{\mathbf{XX}}\|_F^2 \right], \\
		\widehat{\langle \mu_{\mathbb{Q}}, \mu_{\mathbb{Q}} \rangle^2_{\mathcal{H}}} &= \frac{1}{(m)_4} \left[ (\mathbf{1}_m^\top \tilde{\mathbf{K}}_{\mathbf{YY}} \mathbf{1}_m)^2 - 4 \|\tilde{\mathbf{K}}_{\mathbf{YY}} \mathbf{1}_m\|_2^2 + 2 \|\tilde{\mathbf{K}}_{\mathbf{YY}}\|_F^2 \right].
	\end{align}
	
	\item \textbf{mixed products}:
	\begin{align}
		\widehat{\langle \mu_{\mathbb{P}}, \mu_{\mathbb{P}} \rangle_{\mathcal{H}} \langle \mu_{\mathbb{P}}, \mu_{\mathbb{Q}} \rangle_{\mathcal{H}}} &= \frac{1}{m (n)_3} \left[ (\mathbf{1}_n^\top \tilde{\mathbf{K}}_{\mathbf{XX}} \mathbf{1}_n) (\mathbf{1}_n^\top \mathbf{K}_{\mathbf{XY}} \mathbf{1}_m) - 2 \mathbf{1}_n^\top \tilde{\mathbf{K}}_{\mathbf{XX}} \mathbf{K}_{\mathbf{XY}} \mathbf{1}_m \right], \\
		\widehat{\langle \mu_{\mathbb{Q}}, \mu_{\mathbb{Q}} \rangle_{\mathcal{H}} \langle \mu_{\mathbb{Q}}, \mu_{\mathbb{P}} \rangle_{\mathcal{H}}} &= \frac{1}{n (m)_3} \left[ (\mathbf{1}_m^\top \tilde{\mathbf{K}}_{\mathbf{YY}} \mathbf{1}_m) (\mathbf{1}_m^\top \mathbf{K}_{\mathbf{XY}}^\top \mathbf{1}_n) - 2 \mathbf{1}_m^\top \tilde{\mathbf{K}}_{\mathbf{YY}} \mathbf{K}_{\mathbf{XY}}^\top \mathbf{1}_n \right].
	\end{align}
\end{enumerate}

\subsection{Quadratic forms with covariance operators}
\label{B2}

By the definition of the centered covariance operator $C_{\mathbb{P}} = \mathcal{M}_{\mathbb{P}} - \mu_{\mathbb{P}} \otimes \mu_{\mathbb{P}}$, the targeted inner products are constructed by subtracting the unbiased estimates of the corresponding rank-one mean embedding products from the unbiased estimates of the uncentered second-moment projections.

\begin{enumerate}

	\item \textbf{targeted projections}:
	\begin{align}
		\widehat{\langle \mu_{\mathbb{P}}, C_{\mathbb{P}} \mu_{\mathbb{P}} \rangle_{\mathcal{H}}} 
		&= \frac{1}{(n)_3} \sum_i \sum_{j \neq i} \sum_{\ell \notin \{i, j\}} k(\mathbf{x}_i, \mathbf{x}_j) k(\mathbf{x}_i, \mathbf{x}_\ell) - \widehat{\langle \mu_{\mathbb{P}}, \mu_{\mathbb{P}} \rangle^2_{\mathcal{H}}} \nonumber \\
		&= \frac{1}{(n)_3} \left[ \sum_{i,j,\ell} (\tilde{\mathbf{K}}_{\mathbf{XX}})_{ij} (\tilde{\mathbf{K}}_{\mathbf{XX}})_{i\ell} - \sum_{i,j} (\tilde{\mathbf{K}}_{\mathbf{XX}})_{ij}^2 \right] - \widehat{\langle \mu_{\mathbb{P}}, \mu_{\mathbb{P}} \rangle^2_{\mathcal{H}}} \nonumber \\
		&= \frac{1}{(n)_3} \left[ \|\tilde{\mathbf{K}}_{\mathbf{XX}} \mathbf{1}_n\|_2^2 - \|\tilde{\mathbf{K}}_{\mathbf{XX}}\|_F^2 \right] - \widehat{\langle \mu_{\mathbb{P}}, \mu_{\mathbb{P}} \rangle^2_{\mathcal{H}}}, \\[1em]
		\widehat{\langle \mu_{\mathbb{Q}}, C_{\mathbb{Q}} \mu_{\mathbb{Q}} \rangle_{\mathcal{H}}} 
		&= \frac{1}{(m)_3} \sum_j \sum_{i \neq j} \sum_{\ell \notin \{j, i\}} k(\mathbf{y}_j, \mathbf{y}_i) k(\mathbf{y}_j, \mathbf{y}_\ell) - \widehat{\langle \mu_{\mathbb{Q}}, \mu_{\mathbb{Q}} \rangle^2_{\mathcal{H}}} \nonumber \\
		&= \frac{1}{(m)_3} \left[ \sum_{j,i,\ell} (\tilde{\mathbf{K}}_{\mathbf{YY}})_{ji} (\tilde{\mathbf{K}}_{\mathbf{YY}})_{j\ell} - \sum_{j,i} (\tilde{\mathbf{K}}_{\mathbf{YY}})_{ji}^2 \right] - \widehat{\langle \mu_{\mathbb{Q}}, \mu_{\mathbb{Q}} \rangle^2_{\mathcal{H}}} \nonumber \\
		&= \frac{1}{(m)_3} \left[ \|\tilde{\mathbf{K}}_{\mathbf{YY}} \mathbf{1}_m\|_2^2 - \|\tilde{\mathbf{K}}_{\mathbf{YY}}\|_F^2 \right] - \widehat{\langle \mu_{\mathbb{Q}}, \mu_{\mathbb{Q}} \rangle^2_{\mathcal{H}}}.
	\end{align}
	
	\item \textbf{cross-distribution projections}:
	\begin{align}
		\widehat{\langle \mu_{\mathbb{Q}}, C_{\mathbb{P}} \mu_{\mathbb{Q}} \rangle_{\mathcal{H}}} 
		&= \frac{1}{n (m)_2} \sum_i \sum_j \sum_{\ell \neq j} k(\mathbf{x}_i, \mathbf{y}_j) k(\mathbf{x}_i, \mathbf{y}_\ell) - \widehat{\langle \mu_{\mathbb{P}}, \mu_{\mathbb{Q}} \rangle^2_{\mathcal{H}}} \nonumber \\
		&= \frac{1}{n (m)_2} \left[ \sum_{i,j,\ell} (\mathbf{K}_{\mathbf{XY}})_{ij} (\mathbf{K}_{\mathbf{XY}})_{i\ell} - \sum_{i,j} (\mathbf{K}_{\mathbf{XY}})_{ij}^2 \right] - \widehat{\langle \mu_{\mathbb{P}}, \mu_{\mathbb{Q}} \rangle^2_{\mathcal{H}}} \nonumber \\
		&= \frac{1}{n (m)_2} \left[ \|\mathbf{K}_{\mathbf{XY}} \mathbf{1}_m\|_2^2 - \|\mathbf{K}_{\mathbf{XY}}\|_F^2 \right] - \widehat{\langle \mu_{\mathbb{P}}, \mu_{\mathbb{Q}} \rangle^2_{\mathcal{H}}}, \\[1em]
		\widehat{\langle \mu_{\mathbb{P}}, C_{\mathbb{Q}} \mu_{\mathbb{P}} \rangle_{\mathcal{H}}} 
		&= \frac{1}{m (n)_2} \sum_j \sum_i \sum_{\ell \neq i} k(\mathbf{x}_i, \mathbf{y}_j) k(\mathbf{x}_\ell, \mathbf{y}_j) - \widehat{\langle \mu_{\mathbb{P}}, \mu_{\mathbb{Q}} \rangle^2_{\mathcal{H}}} \nonumber \\
		&= \frac{1}{m (n)_2} \left[ \sum_{j,i,\ell} (\mathbf{K}_{\mathbf{XY}}^\top)_{ji} (\mathbf{K}_{\mathbf{XY}}^\top)_{j\ell} - \sum_{j,i} (\mathbf{K}_{\mathbf{XY}}^\top)_{ji}^2 \right] - \widehat{\langle \mu_{\mathbb{P}}, \mu_{\mathbb{Q}} \rangle^2_{\mathcal{H}}} \nonumber \\
		&= \frac{1}{m (n)_2} \left[ \|\mathbf{K}_{\mathbf{XY}}^\top \mathbf{1}_n\|_2^2 - \|\mathbf{K}_{\mathbf{XY}}\|_F^2 \right] - \widehat{\langle \mu_{\mathbb{P}}, \mu_{\mathbb{Q}} \rangle^2_{\mathcal{H}}}.
	\end{align}
	
	\item \textbf{mixed projections}:
	\begin{align}
		\widehat{\langle \mu_{\mathbb{P}}, C_{\mathbb{P}} \mu_{\mathbb{Q}} \rangle_{\mathcal{H}}} 
		&= \frac{1}{m (n)_2} \sum_i \sum_{j \neq i} \sum_\ell k(\mathbf{x}_i, \mathbf{x}_j) k(\mathbf{x}_i, \mathbf{y}_\ell) - \widehat{\langle \mu_{\mathbb{P}}, \mu_{\mathbb{P}} \rangle_{\mathcal{H}} \langle \mu_{\mathbb{P}}, \mu_{\mathbb{Q}} \rangle_{\mathcal{H}}} \nonumber \\
		&= \frac{1}{m (n)_2} \sum_{i,j,\ell} (\tilde{\mathbf{K}}_{\mathbf{XX}})_{ij} (\mathbf{K}_{\mathbf{XY}})_{i\ell} - \widehat{\langle \mu_{\mathbb{P}}, \mu_{\mathbb{P}} \rangle_{\mathcal{H}} \langle \mu_{\mathbb{P}}, \mu_{\mathbb{Q}} \rangle_{\mathcal{H}}} \nonumber \\
		&= \frac{1}{m (n)_2} \mathbf{1}_n^\top \tilde{\mathbf{K}}_{\mathbf{XX}} \mathbf{K}_{\mathbf{XY}} \mathbf{1}_m - \widehat{\langle \mu_{\mathbb{P}}, \mu_{\mathbb{P}} \rangle_{\mathcal{H}} \langle \mu_{\mathbb{P}}, \mu_{\mathbb{Q}} \rangle_{\mathcal{H}}}, \\[1em]
		\widehat{\langle \mu_{\mathbb{Q}}, C_{\mathbb{Q}} \mu_{\mathbb{P}} \rangle_{\mathcal{H}}} 
		&= \frac{1}{n (m)_2} \sum_j \sum_{i \neq j} \sum_\ell k(\mathbf{y}_j, \mathbf{y}_i) k(\mathbf{y}_j, \mathbf{x}_\ell) - \widehat{\langle \mu_{\mathbb{Q}}, \mu_{\mathbb{Q}} \rangle_{\mathcal{H}} \langle \mu_{\mathbb{Q}}, \mu_{\mathbb{P}} \rangle_{\mathcal{H}}} \nonumber \\
		&= \frac{1}{n (m)_2} \sum_{j,i,\ell} (\tilde{\mathbf{K}}_{\mathbf{YY}})_{ji} (\mathbf{K}_{\mathbf{XY}}^\top)_{j\ell} - \widehat{\langle \mu_{\mathbb{Q}}, \mu_{\mathbb{Q}} \rangle_{\mathcal{H}} \langle \mu_{\mathbb{Q}}, \mu_{\mathbb{P}} \rangle_{\mathcal{H}}} \nonumber \\
		&= \frac{1}{n (m)_2} \mathbf{1}_m^\top \tilde{\mathbf{K}}_{\mathbf{YY}} \mathbf{K}_{\mathbf{XY}}^\top \mathbf{1}_n - \widehat{\langle \mu_{\mathbb{Q}}, \mu_{\mathbb{Q}} \rangle_{\mathcal{H}} \langle \mu_{\mathbb{Q}}, \mu_{\mathbb{P}} \rangle_{\mathcal{H}}}.
	\end{align}
\end{enumerate}

\subsection{Squared kernel terms}
\begin{align}
	\widehat{\mathbb{E}\left[ k(X, X')^2 \right]} &= \frac{1}{(n)_2} \|\tilde{\mathbf{K}}_{\mathbf{XX}}\|_F^2, \\
	\widehat{\mathbb{E}\left[ k(Y, Y')^2 \right]} &= \frac{1}{(m)_2} \|\tilde{\mathbf{K}}_{\mathbf{YY}}\|_F^2, \\
	\widehat{\mathbb{E}\left[ k(X, Y)^2 \right]} &= \frac{1}{nm} \|\mathbf{K}_{\mathbf{XY}}\|_F^2.
\end{align}

\section{Random Fourier Features (RFF) approximation}
\label{app3}
Based on Bochner's theorem \cite{rahimi2007random}, a shift-invariant kernel $k(x, y)$ can be approximated by the inner product of randomized feature maps. By drawing $L$ independent random weights $\mathbf{W} \in \mathbb{R}^{d \times L}$ from the corresponding Fourier transform distribution (e.g., standard normal for the Gaussian kernel, or Cauchy for the Laplace kernel), we construct the low-dimensional feature matrices $\mathbf{Z}_{\mathbf{X}} \in \mathbb{R}^{n \times 2L}$ and $\mathbf{Z}_{\mathbf{Y}} \in \mathbb{R}^{m \times 2L}$:
\begin{align}
	\mathbf{Z}_{\mathbf{X}} &= \frac{1}{\sqrt{L}} \big[ \cos(\mathbf{X}\mathbf{W}), \sin(\mathbf{X}\mathbf{W}) \big], \\
	\mathbf{Z}_{\mathbf{Y}} &= \frac{1}{\sqrt{L}} \big[ \cos(\mathbf{Y}\mathbf{W}), \sin(\mathbf{Y}\mathbf{W}) \big].
\end{align}
The kernel matrices can thus be explicitly approximated as low-rank inner products: $\mathbf{K}_{\mathbf{XX}} \approx \mathbf{Z}_{\mathbf{X}}\mathbf{Z}_{\mathbf{X}}^\top$, $\mathbf{K}_{\mathbf{YY}} \approx \mathbf{Z}_{\mathbf{Y}}\mathbf{Z}_{\mathbf{Y}}^\top$, and $\mathbf{K}_{\mathbf{XY}} \approx \mathbf{Z}_{\mathbf{X}}\mathbf{Z}_{\mathbf{Y}}^\top$.

To compute the unbiased variance estimators, we must meticulously handle the diagonal elements (hollow matrices). Let $\mathbf{s}_X = \mathbf{Z}_{\mathbf{X}}^\top \mathbf{1}_n \in \mathbb{R}^{2L}$ and $\mathbf{s}_Y = \mathbf{Z}_{\mathbf{Y}}^\top \mathbf{1}_m \in \mathbb{R}^{2L}$ denote the column sum vectors in the feature space. We also extract the diagonal vectors $\mathbf{d}_X = \mathrm{diag}(\mathbf{Z}_{\mathbf{X}}\mathbf{Z}_{\mathbf{X}}^\top) \in \mathbb{R}^n$ and $\mathbf{d}_Y = \mathrm{diag}(\mathbf{Z}_{\mathbf{Y}}\mathbf{Z}_{\mathbf{Y}}^\top) \in \mathbb{R}^m$. By exploiting the associativity of matrix multiplication, we reformulate the costly matrix-vector operations as follows.

\paragraph{Single-sample kernel matrix}
For the $\mathbf{X}$ single-sample hollow kernel matrix $\tilde{\mathbf{K}}_{\mathbf{XX}}$, let $\mathbf{C}_{\mathbf{XX}} = \mathbf{Z}_{\mathbf{X}}^\top \mathbf{Z}_{\mathbf{X}} \in \mathbb{R}^{2L \times 2L}$. We eliminate the $\mathcal{O}(n^2)$ dependency by rewriting the terms as
\begin{align}
	\|\tilde{\mathbf{K}}_{\mathbf{XX}}\|_F^2 &= \|\mathbf{C}_{\mathbf{XX}}\|_F^2 - \|\mathbf{d}_X\|_2^2, \\
	\|\tilde{\mathbf{K}}_{\mathbf{XX}}\mathbf{1}_n\|_2^2 &= \|\mathbf{Z}_{\mathbf{X}} \mathbf{s}_X - \mathbf{d}_X\|_2^2, \\
	(\mathbf{1}_n^\top \tilde{\mathbf{K}}_{\mathbf{XX}}\mathbf{1}_n)^2 &= \big(\|\mathbf{s}_X\|_2^2 - \mathbf{1}_n^\top \mathbf{d}_X\big)^2.
\end{align}
The corresponding terms for the $\mathbf{Y}$ single-sample kernel matrix $\tilde{\mathbf{K}}_{\mathbf{YY}}$ follow identically by substituting $\mathbf{Z}_{\mathbf{Y}}$, $\mathbf{s}_Y$, $\mathbf{d}_Y$, and $\mathbf{C}_{\mathbf{YY}} = \mathbf{Z}_{\mathbf{Y}}^\top \mathbf{Z}_{\mathbf{Y}}$.

\paragraph{Cross-sample kernel matrix}
For the cross-sample matrix $\mathbf{K}_{\mathbf{XY}}$, we apply the trace trick to compute the Frobenius norm without explicitly materializing the $n \times m$ matrix. Since $\|\mathbf{K}_{\mathbf{XY}}\|_F^2 = \mathrm{Tr}(\mathbf{K}_{\mathbf{XY}}^\top \mathbf{K}_{\mathbf{XY}}) \approx \mathrm{Tr}(\mathbf{Z}_{\mathbf{Y}} \mathbf{Z}_{\mathbf{X}}^\top \mathbf{Z}_{\mathbf{X}} \mathbf{Z}_{\mathbf{Y}}^\top) = \mathrm{Tr}(\mathbf{C}_{\mathbf{XX}} \mathbf{C}_{\mathbf{YY}})$, the cross-accumulators become:
\begin{align}
	\|\mathbf{K}_{\mathbf{XY}}\|_F^2 &= \mathrm{Tr}(\mathbf{C}_{\mathbf{XX}} \mathbf{C}_{\mathbf{YY}}), \\
	\|\mathbf{K}_{\mathbf{XY}}\mathbf{1}_m\|_2^2 &= \|\mathbf{Z}_{\mathbf{X}} \mathbf{s}_Y\|_2^2, \\
	\|\mathbf{K}_{\mathbf{XY}}^\top \mathbf{1}_n\|_2^2 &= \|\mathbf{Z}_{\mathbf{Y}} \mathbf{s}_X\|_2^2, \\
	(\mathbf{1}_n^\top \mathbf{K}_{\mathbf{XY}} \mathbf{1}_m)^2 &= (\mathbf{s}_X^\top \mathbf{s}_Y)^2.
\end{align}

\paragraph{Mixed cross-covariance projections}
Crucially, evaluating the first-order variance components requires the mixed cross-covariance projections. By factoring the feature matrices, these terms are efficiently obtained via linear-time dot products:
\begin{align}
	\mathbf{1}_n^\top \tilde{\mathbf{K}}_{\mathbf{XX}} \mathbf{K}_{\mathbf{XY}} \mathbf{1}_m &= (\mathbf{Z}_{\mathbf{X}} \mathbf{s}_X - \mathbf{d}_X)^\top (\mathbf{Z}_{\mathbf{X}} \mathbf{s}_Y), \\
	\mathbf{1}_m^\top \tilde{\mathbf{K}}_{\mathbf{YY}} \mathbf{K}_{\mathbf{XY}}^\top \mathbf{1}_n &= (\mathbf{Z}_{\mathbf{Y}} \mathbf{s}_Y - \mathbf{d}_Y)^\top (\mathbf{Z}_{\mathbf{Y}} \mathbf{s}_X).
\end{align}

Based on this theoretical breakdown, the FastMMD framework presents a complete accelerated procedure for computing the unbiased MMD along with its exact variance estimator. By operating entirely within the $2L$-dimensional feature space and utilizing the trace trick for cross-Frobenius norms, the algorithm seamlessly reduces the overall time complexity from the naive $\mathcal{O}(N^2 d)$ to $\mathcal{O}(N L d + N L^2)$. Furthermore, it simultaneously slashes the space complexity from a prohibitive $\mathcal{O}(N^2)$ down to $\mathcal{O}(N L)$, where $d$ denotes the original feature dimension.

\end{appendices}


\bibliography{references}% common bib file

\end{document}